\documentclass{IEEEtran}
\usepackage{cite}
\usepackage{amsmath,amssymb,amsfonts}
\usepackage{algorithmic}
\usepackage{graphicx}
\usepackage{textcomp}
\usepackage{multirow}
\usepackage{hyperref}
\usepackage{multicol}
\usepackage{url}
\usepackage{booktabs}
\usepackage{makecell}
\usepackage{color}
\usepackage{bbding}
\newcommand{\tabincell}[2]{\begin{tabular}{@{}#1@{}}#2\end{tabular}}

\def\BibTeX{{\rm B\kern-.05em{\sc i\kern-.025em b}\kern-.08em
		T\kern-.1667em\lower.7ex\hbox{E}\kern-.125emX}}
\ifCLASSINFOpdf
\else
\fi
\begin{document}
      \title{Real-Time Multi-Scene Visibility Enhancement for Promoting Navigational Safety of Vessels Under Complex Weather Conditions}
      \author{Ryan Wen Liu, \textit{Member, IEEE},
            Yuxu Lu,
            Yuan Gao,
            Yu Guo,
            Wenqi Ren, \textit{Member, IEEE},\\
            Fenghua Zhu, \textit{Senior Member, IEEE},
            and Fei-Yue Wang, \textit{Fellow, IEEE}
		%
            %
		%\thanks{This work was supported in part by the National Natural Science Foundation of China under Grants 52422111 and 52271365, and in part by the Excellent Youth Foundation of Hubei Scientific Committee under Grant 2024AFA042. \textit{(Corresponding authors: Yuxu Lu; Fenghua Zhu.)}}
		%
        \thanks{Ryan Wen Liu, Yuan Gao, and Yu Guo are with the School of Navigation, Wuhan University of Technology, Wuhan 430063, China, and also with the State Key Laboratory of Maritime Technology and Safety, Wuhan, 430063, China (e-mail: \{wenliu, yuangao, yuguo\}@whut.edu.cn).}
		\thanks{Yuxu Lu is with the Department of Logistics and Maritime Studies, Hong Kong Polytechnic University, Hong Kong (e-mail: yuxulouis.lu@connect.polyu.hk).}
		\thanks{Wenqi Ren is with the School of Cyber Science and Technology, Sun Yat-sen University at Shenzhen, Shenzhen 518107, China (e-mail: renwq3@mail.sysu.edu.cn).}
		\thanks{Fenghua Zhu and Fei-Yue Wang are with the State Key Laboratory of Multimodal Artificial Intelligence Systems, Institute of Automation, Chinese Academy of Sciences, Beijing 100190, China (e-mail: \{fenghua.zhu, feiyue.wang\}@ia.ac.cn).}
		
	}
	
	\maketitle
	
	\begin{abstract}
        The visible-light camera, which is capable of environment perception and navigation assistance, has emerged as an essential imaging sensor for marine surface vessels in intelligent waterborne transportation systems (IWTS). However, the visual imaging quality inevitably suffers from several kinds of degradations (e.g., limited visibility, low contrast, color distortion, etc.) under complex weather conditions (e.g., haze, rain, and low-lightness). The degraded visual information will accordingly result in inaccurate environment perception and delayed operations for navigational risk. To promote the navigational safety of vessels, many computational methods have been presented to perform visual quality enhancement under poor weather conditions. However, most of these methods are essentially \textit{specific-purpose} implementation strategies, only available for one specific weather type. To overcome this limitation, we propose to develop a \textit{general-purpose} multi-scene visibility enhancement method, i.e., edge reparameterization- and attention-guided neural network (ERANet), to adaptively restore the degraded images captured under different weather conditions. In particular, our ERANet simultaneously exploits the channel attention, spatial attention, and reparameterization technology to enhance the visual quality while maintaining low computational cost. Extensive experiments conducted on standard and IWTS-related datasets have demonstrated that our ERANet could outperform several representative visibility enhancement methods in terms of both imaging quality and computational efficiency. The superior performance of IWTS-related object detection and scene segmentation could also be steadily obtained after ERANet-based visibility enhancement under complex weather conditions.
	\end{abstract}
	
	
	\section{Introduction}
	\label{sec:introduction}
\begin{figure}[t]
        \centering
        \setlength{\abovecaptionskip}{0.cm}
        \includegraphics[width=1.00\linewidth]{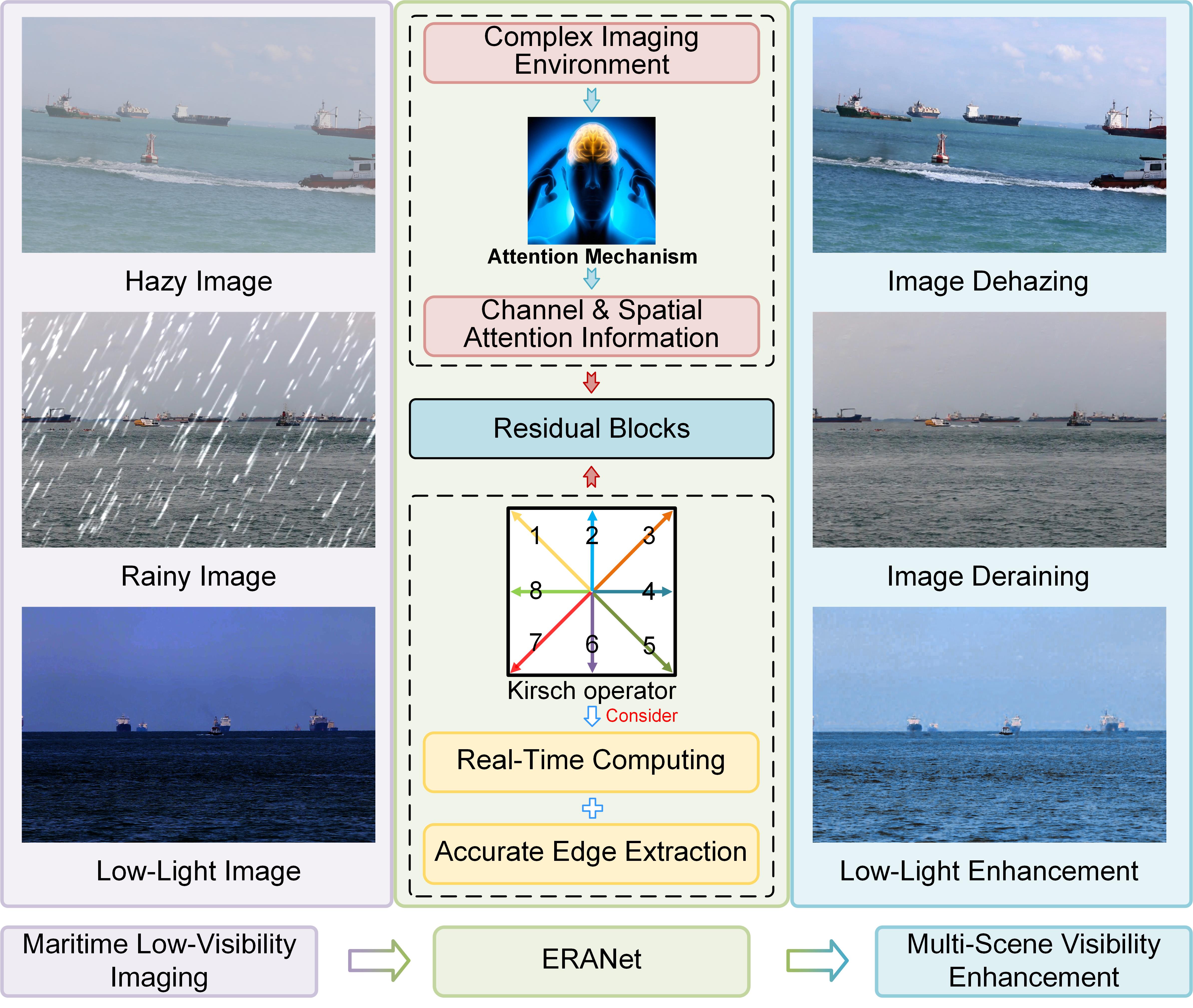}
        \caption{The workflow of ERANet-based real-time multi-scene low-visibility scene recovery for intelligent waterborne transportation systems (IWTS).}
        \label{Figure_01}
    \end{figure}

	\IEEEPARstart{T}{he} visible-light camera has emerged as an essential imaging sensor for both manned and unmanned surface vessels in intelligent waterborne transportation systems (IWTS) \cite{german2012long,thombre2020sensors}. The onboard visible-light imaging devices are highly dependent upon the weather conditions. As shown in Fig. \ref{Figure_01}, the captured images easily suffer from obvious degradations (e.g., limited visibility, low contrast, and color distortion) under different weather or light conditions (e.g., haze, rain, and low-lightness). The quality-degraded images will bring negative effects on traffic situational awareness and navigation safety, etc \cite{liu2021enhanced,zhang2022trans4trans}. Moreover, the low-quality images often yield even more significant difficulties in multi-sensor data fusion \cite{liu2022intelligent}, image compression and reconstruction, particularly for low-resource edge computing devices \cite{bairi2022pscs}. It is thus necessary to improve the visual image quality to promote the navigational safety of moving vessels under poor weather conditions.

    In the literature, the traditional model-based and advanced learning-based computational methods have been exploited to enhance the visual perception of common low-visibility scenes. For example, in the case of haze weather, the representative dark channel prior (DCP) \cite{he2010single} and other popular priors, e.g., non-local prior (NLP) \cite{berman2018single}, saturation line prior (SLP) \cite{ling2023single}, and region line prior (RLP) \cite{ju2021idrlp}, etc., have contributed to the model-based dehazing methods. Due to the strong representation capacity of deep learning, the convolutional neural network (CNN) \cite{liu2022deep}, generative adversarial network (GAN) \cite{wang2022cycle}, Transformer network \cite{valanarasu2022transweather,wang2023uscformer}, and their extensions \cite{sahu2022novel} have significantly promoted the visual imaging quality. The current model-based deraining methods mainly include image decomposition and filtering strategies \cite{yang2020single}. To preserve essential structures during deraining, both CNN and GAN have also been exploited to directly remove the rain streaks and raindrops from the degraded images \cite{kulkarni2022wipernet,zhang2023data}. The low-lightness is a more common weather phenomenon in different modes of transport. The low-light image enhancement has thus attracted huge attention from both academia and industry \cite{liu2021benchmarking}. The Retinex and its extensions \cite{wang2013naturalness,liu2023low} have become the representative model-based low-visibility enhancement methods. The recent studies on deep learning-based low-light image enhancement can be found in reviews \cite{liu2021benchmarking,wang2022low} and references therein. However, most of these methods are essentially \textit{specific-purpose} visibility enhancement strategies, only available for one specific weather type. The flexibility and applicability of these imaging methods will be inevitably degraded under different weather conditions.
	\begin{figure*}[t]
		\centering
		\includegraphics[width=1.00\linewidth]{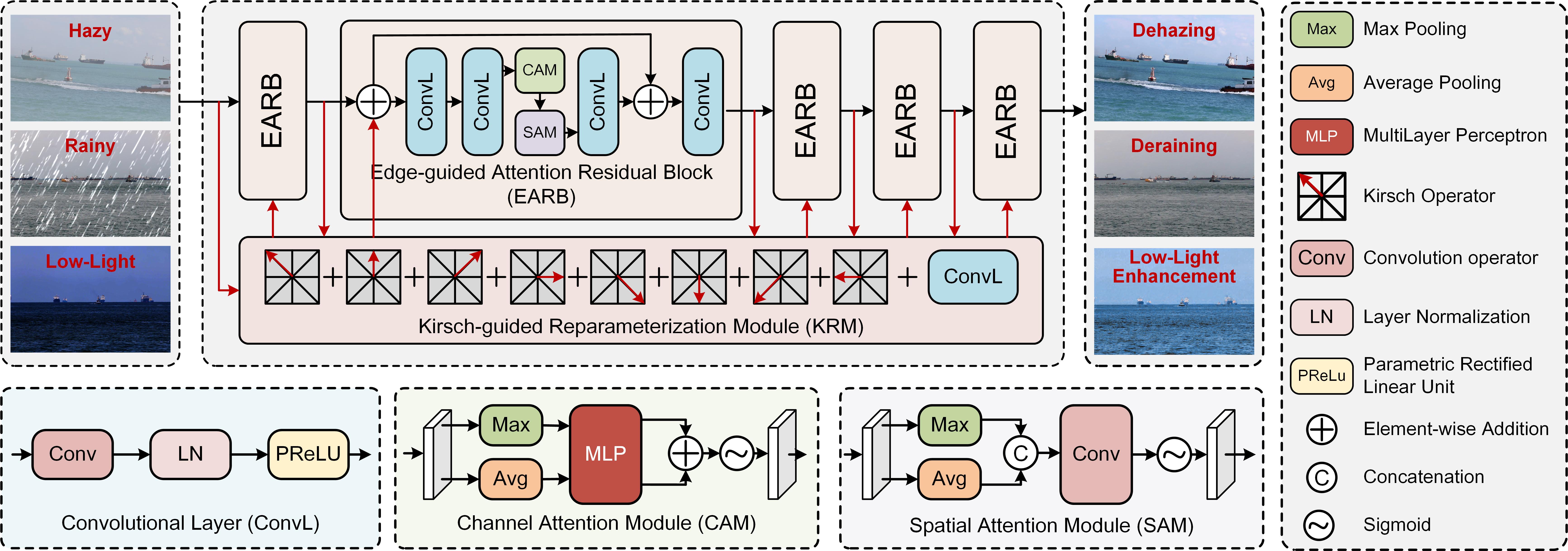}
		\caption{The flowchart of our edge reparameterization- and attention-guided network (ERANet) for multi-scene visibility enhancement (i.e., hazy, rainy, and low-light). The applications of channel attention, spatial attention, and Kirsch eight-directions-guided edge parameterization are able to balance the low-visibility scene restoration and computational cost.}
		\label{Figure02}
	\end{figure*}

    To make one visibility enhancement method available for different types of weather, many \textit{general-purpose} multi-scene visibility enhancement strategies have been recently introduced in the literature. For example, Liu \textit{et al.} \cite{liu2022rank} propose a rank-one prior (ROP)-based physical imaging method to restore degraded images under different weather conditions, such as sand dust, haze, and low-light. The denoising diffusion probabilistic models (DDPM) have been exploited to effectively implement vision restoration under snowy, rainy, and hazy conditions \cite{ozdenizci2023restoring}. By decoupling the degradation and background features, Cheng \textit{et al}. \cite{cheng2023deep} propose a deep fuzzy clustering Transformer (DFCFormer) for performing multi-task image restoration under rainy and hazy conditions. This work mainly considers the hazy, rainy, and low-light conditions, which are more common for marine surface vessels in IWTS. The current advanced imaging methods could not be directly employed to perform multi-task image restoration under these weather conditions. Therefore, we propose to develop an edge reparameterization- and attention-guided neural network (ERANet) to recover high-quality images from various image degradations. In particular, our method can extract the gradient information in eight directions through the Kirsch operators, which could be reparameterized into a single convolutional layer. We then adopt the channel and spatial attention mechanisms to enhance learning and map color and spatial edge features. Our ERANet can generate real-time recovery of maritime low-visibility scenes while minimizing the additional computational parameters. The main contributions of this work are summarized as follows
    \begin{itemize}
        \item An edge reparameterization- and attention-guided neural network (ERANet), i.e., a \textit{general-purpose} multi-scene visibility enhancement method, is proposed to adaptively reconstruct the quality-degraded maritime images captured under complex weather conditions.
        \item A reparameterization module, which exploits the self-defined Kirsch gradient operators with eight directions, is proposed to effectively extract the meaningful edge features from the low-visibility images. It is beneficial for ERANet to suppress unwanted outliers and preserve important image structures.
        \item Both quantitative and qualitative experiments demonstrate that our \textit{general-purpose} ERANet can significantly improve the visual image quality in different weather conditions. In addition, it could prove to be efficient with lower computational cost than state-of-the-art methods, which has significant industrial application value for manned and unmanned surface vessels in IWTS.
    \end{itemize}

	The remainder of this work is organized as follows. The current studies on dehazing, deraining, low-light enhancement, and multi-scene visibility enhancement are reviewed in Section \ref{relatedwork}. Our multi-scene visibility enhancement method is detailedly described in Section \ref{earnet}. Experimental results and discussion are provided in Section \ref{exper}. Section \ref{conc} finally presents the conclusions and future perspectives.
\section{Related Work}
\label{relatedwork}
	In this section, we will briefly review the recent studies on low-visibility enhancement in different weather conditions.
\subsection{Dehazing}
    The physical imaging model of real-world hazy images can be defined as follows
    \begin{equation}
        \mathbf{I}_{h}(x)=\mathbf{J}(x)\mathbf{t}(x) + \mathbf{A}(1-\mathbf{t}(x)),
        \label{Equation_atmospheric scattering model}
    \end{equation}
    where $x$ denotes the pixel index of the image, $\mathbf{J}$ is the scene radiance, $\mathbf{t}$ is the medium transmission relative to the depth of the scene, $\mathbf{A}$ is the global atmospheric light, and $\mathbf{I}_{h}$ is the degraded hazy image.
    
    Dehazing methods mainly include physical priors- \cite{he2010single,liu2022rank} and learning-based \cite{ren2016single,li2017aod,chen2019gated,liu2022deep}. DCP \cite{he2010single} reveals the general statistical laws of hazy images, upon which many improved methods \cite{peng2018generalization} have been proposed. However, in the complex imaging environment, DCP-guided methods are not fully applicable, and even the image after dehazing has color distortion in bright areas (such as the sky and water surface) \cite{liu2022deep}. Learning-based methods can be better applied to complex hazy imaging. Earlier CNN-based dehazing methods estimate the transmission of atmospheric scattering models (such as MSCNN \cite{ren2016single}) to reconstruct haze-free images. Li \textit{et al}. \cite{li2017aod} propose a reformulated atmospheric scattering model to reduce the generation of unsatisfactory dehazed images caused by inaccurate estimates of transmittance and atmospheric light value parameters. The generative adversarial network (GAN) \cite{wang2022cycle} and Transformer \cite{guo2022image} have also been successfully applied in complex dehazing tasks.
\subsection{Deraining}
   The widely-used degradation model, which expresses the input rainy image, is formulated as follows
    \begin{equation}
        \mathbf{I}_{r}(x)=\mathbf{O}(x) + \mathbf{S}(x),
        \label{Equation_rain model}
    \end{equation}
    where $\mathbf{O}$ is the rain-free background scene, $\mathbf{S}$ is the rain streak layer, and $\mathbf{I}_{r}$ is the captured rainy image.
    
    The image deraining methods can be broadly divided into two categories, i.e., model/prior- and learning-based methods. The model/prior-based methods mainly analyze the high and low-frequency information of rainy images and separate rain streaks and background details through sparse representation \cite{kang2012self}, guided filtering \cite{xu2012removing}, Gaussian mixture models \cite{li2016rain}, etc. However, the complex optimization process of traditional methods will bring additional calculation time costs. The learning-based methods can make a balance between restoration performance and computational time. In addition, valuable constraints and prior information can be optimized further to improve the estimation of potential rain-free images \cite{fu2017clearing,wang2020model}. Rain removal requires an accurate extraction of rain streaks' structure and motion information. To improve the network's generalization capability, many efforts have been dedicated towards rain removal using the fully supervised, semi-supervised \cite{wei2021semi}, and unsupervised \cite{ye2022unsupervised} learning methods. Benefiting from the strong learning capacities, these methods have significantly improved the deraining performance.
\subsection{Low-Light Image Enhancement}
    Land \textit{et al}. \cite{land1977retinex} propose the Retinex theory, which assumes that an image $\mathbf{I}_{l}$ can be decomposed into illumination $\mathbf{L}$ and reflection $\mathbf{R}$, to model the low-light image, i.e.,
    \begin{equation}
	\mathbf{I}_{l}(x)= \mathbf{L}(x) * \mathbf{R}(x),
        \label{Equation_low-light model}
    \end{equation}
where the reflection $\mathbf{R}$ contains rich color, texture, and detail information. In contrast, the illumination $\mathbf{L}$ only includes brightness smoothly distributed in the image domain.

The Retinex-based methods \cite{wang2013naturalness} can accurately extract the normal-light images in simple low-light scenes. However, in complex imaging scenes, the inaccurate reflection components would cause the generated images to appear unnatural effects in brightness and color. The histogram equalization (HE)- \cite{pizer1987adaptive} and dehazing-based \cite{jiang2013night} methods provide powerful solutions for low-light image enhancement as well. The learning-based methods \cite{jiang2022unsupervised} have a stronger feature extraction ability and can more accurately extract normal light information from dark backgrounds. The model-driven learning methods \cite{wei2018deep,zhang2019kindling} can further improve the robustness and stability.

\subsection{Multi-Scene Visibility Enhancement}
   The imaging environment in the real world is unpredictable. Therefore, researchers have proposed various visual perception enhancement methods in harsh imaging scenarios under the unified framework of model- and learning-based. Liu \textit{et al}. \cite{liu2022rank} propose a rank-one prior (ROP) to optimize the estimate of transmittance to real-timely restore different scene degradation images. Sindagi \textit{et al}. \cite{sindagi2020prior} propose an unsupervised prior-based domain-adversarial object detection framework, which improves the recognition accuracy of the detector in hazy and rainy conditions. Zamir \textit{et al}. \cite{zamir2021multi} adopt a multi-stage architecture to achieve a complex balance between spatial details and high-level contextual information when restoring images. Zhou \textit{et al}. \cite{zhou2023fourmer} utilize the Fourier transform to separate image degradation and content, enabling global modeling and achieving competitive performance with fewer computational resources. The ProRes \cite{ma2023prores} is essentially a Transformer-based universal imaging framework, which proposes degradation-aware visual prompts for several different image restoration tasks, such as denoising, deraining, low-light enhancement, and deblurring. Gao \textit{et al}. \cite{gao2023prompt} propose a data ingredient-oriented method, which combines prompt-based learning, CNNs, Transformers, and a feature fusion mechanism, to efficiently handle an extensive range of image degradation tasks with reduced computational requirements. The AirNet \cite{li2022all}, IDR \cite{zhang2023ingredient}, and DFCFormer \cite{cheng2023deep} can recover images from various unknown types and levels of corruption with a single trained model. The denoising diffusion probability model (DDPM)-based method \cite{ozdenizci2023restoring} is successfully applied in the scene restoration task, but a huge amount of computation inevitably accompanies it. 
\section{The Proposed Network}\label{earnet}
   The rapid development of IWTS places greater demands on the image data collected by visual sensors. The haze, rain, and low-lightness are common low-visibility scenarios in maritime practice. Many efforts have been devoted to performing low-visibility enhancement under these scenarios. Due to the insufficient computing power of edge devices for manned and unmanned surface vehicles, it is necessary to develop a general lightweight low-visibility enhancement network. As shown in Fig. \ref{Figure02}, we propose an edge-reparameterization and attention-guided network (ERANet), which achieves multi-scene visibility enhancement through a single network, in this work. It primarily consists of four parts, i.e., a normal convolutional layer (ConvL), a Kirsch-guided reparameterization module (KRM), a channel attention module (CAM), and a spatial attention module (SAM).
	\setlength{\tabcolsep}{5.0pt}
	\begin{table*}[t]
		\centering
		
		\caption{Self-defined gradient detection operators in eight directions of Kirsch.}
		\begin{tabular}{c|cccccccc}
			\hline
			& $K_1$ & $K_2$  & $K_3$ & $K_4$ & $K_5$ & $K_6$ & $K_7$ & $K_8$ \\ \hline\hline
			Direction & $\nwarrow$ & $\uparrow$ & $\nearrow$ & $\rightarrow$ & $\searrow$ & $\downarrow$ & $\swarrow$ & $\leftarrow$ \\ \hline
			Operator  & \begin{tabular}[c]{@{}c@{}}$\left[\begin{smallmatrix}\\      +5 & +5 & -3 \\\\      +5 & 0 & -3 \\\\      -3 & -3 & -3\\      \end{smallmatrix}\right]$\end{tabular} & \begin{tabular}[c]{@{}c@{}}$\left[\begin{smallmatrix}\\      +5 & +5 & +5 \\\\      -3 & 0 & -3 \\\\      -3 & -3 & -3\\      \end{smallmatrix}\right]$\end{tabular} & \begin{tabular}[c]{@{}c@{}}$\left[\begin{smallmatrix}\\      -3 & +5 & +5 \\\\      -3 & 0 & +5 \\\\      -3 & -3 & -3\\      \end{smallmatrix}\right]$\end{tabular} & \begin{tabular}[c]{@{}c@{}}$\left[\begin{smallmatrix}\\      -3 & -3 & +5 \\\\      -3 & 0 & +5 \\\\      -3 & -3 & +5\\      \end{smallmatrix}\right]$\end{tabular} & \begin{tabular}[c]{@{}c@{}}$\left[\begin{smallmatrix}\\      -3 & -3 & -3 \\\\      -3 & 0 & +5 \\\\      -3 & +5 & +5\\      \end{smallmatrix}\right]$\end{tabular} & \begin{tabular}[c]{@{}c@{}}$\left[\begin{smallmatrix}\\      -3 & -3 & -3 \\\\      -3 & 0 & -3 \\\\      +5 & +5 & +5\\      \end{smallmatrix}\right]$\end{tabular} & \begin{tabular}[c]{@{}c@{}}$\left[\begin{smallmatrix}\\      -3 & -3 & -3 \\\\      +5 & 0 & -3 \\\\      +5 & +5 & -3\\      \end{smallmatrix}\right]$\end{tabular} & \begin{tabular}[c]{@{}c@{}}$\left[\begin{smallmatrix}\\      +5 & -3 & -3 \\\\      +5 & 0 & -3 \\\\      +5 & -3 & -3\\      \end{smallmatrix}\right]$\end{tabular} \\ \hline
		\end{tabular}\label{Table_Kirsch}
	\end{table*}

\subsection{Convolutional Layer}
    In this work, the normal convolutional layer is exploited for network parameter learning and mapping. To better adapt to the image restoration tasks under complex weather conditions, the layer normalization ($\operatorname{LN}$) \cite{ba2016layer} is utilized to optimize the global features in each channel during network learning process. In this work, the normal convolutional layer (ConvL) is defined as follows
    \begin{equation}\label{eq:convolutional}
        \mathrm{ConvL}(x_{in})=\operatorname{PR}(\mathrm{LN}(\mathrm{Conv}(x_{in}))),
    \end{equation}
    where $x_{in} \in \mathbb{R}^{C \times H \times W}$ is the input of $\mathrm{{ConvL}}$, $\mathrm{Conv}$ and $\mathrm{PR}$ represent the convolution operation and parametric rectified linear unit (PReLU), respectively. To balance the imaging performance and computational cost, the number of channels for each $\mathrm{ConvL}$ is set to $32$. We find that the imaging results for this manually-selected parameter are consistently promising under different severe weather conditions.
\subsection{Attention Mechanism}
    The attention mechanism can realize the efficient allocation of information processing resources and can give more attention to key scenes while temporarily ignoring the unimportant scenes \cite{woo2018cbam}. The maritime scenes are mainly composed of sky and water regions. Although vessels occupy a small area of the entire image, it is what we should focus on when enhancing maritime images. Therefore, the attention mechanism is used to focus on important information with high weights, ignore irrelevant information with low weights, and continuously adjust the weights during the network learning process. Therefore, a single model can extract more valuable feature information in different imaging environments. As shown in Fig. \ref{Figure02}, both channel attention and spatial attention are jointly exploited to further improve the scene restoration performance.
\subsubsection{Channel Attention Module}
    The correlation of histogram distribution among three channels of the image collected in hazy and low-light scenes is commonly weakened. Therefore, the visibility will be insufficient, and the target features will be unobvious. The channel attention can assist the model in discerning distinctions in color, form, and other attributes of water targets that are deteriorating due to adverse weather. This information could potentially be distributed across different channels within the image. The channel attention mechanism is thus exploited to reconstruct the relationship between feature channels and correct the incorrect colors of low-visibility images. Woo \textit{et al.} \cite{woo2018cbam} propose to aggregate and collect the spatial information of channels by average pooling ($\mathrm{Avg}$) and max pooling ($\operatorname{Max}$). To strengthen the inter-channel correlation during the learning process, the spatial dimension of the hidden layer feature map will be compressed. After being used for learning and mapping by multilayer perceptron with shared parameters, the spatial dimension of the feature map will be restored. The channel attention module ($\mathrm{CAM}$) can thus be defined as follows
	\begin{equation}
		\begin{aligned}
			\operatorname{CAM}
			&=\sigma(\mathrm{MLP}(\mathrm{Avg}(x_{in}^c)) + \mathrm{MLP}(\mathrm{Max}(x_{in}^c)))\\
			&=\sigma\left(\mathrm{MLP}\left(F_{avg}^{c}\right) + \mathrm{MLP}\left((F_{max}^{c}\right)\right),
		\end{aligned}
	\end{equation}
	where $x_{in}^c$ is the input of $\operatorname{CAM}$, $\operatorname{MLP}$ denotes the multilayer perceptron, $\sigma$ is the Sigmoid nonlinear activation function, and $F_{avg}^{c}$ and $F_{max}^{c}$ denote the average-pooled and max-pooled features, respectively.
\subsubsection{Spatial Attention Module}
    The spatial attention can assist the model in concentrating on crucial image regions. The information, such as the location of water targets and the direction of unwanted rain lines, could be reflected in the spatial distribution of the image. The high-frequency information, such as unwanted rain streaks in the rainy images, and obscured edge features in the hazy and low-light images, would get more attention through the spatial attention module. We apply the average pooling and max pooling operations along the channel axis and generate average pooled features $F_{avg}^{s} \in \mathbb{R}^{1 \times H \times W}$ and the max pooled features $F_{max}^{s} \in \mathbb{R}^{1 \times H \times W}$ in the channel. The standard convolution $\mathrm{Conv}^{7 \times 7}$ with a kernel of $7$ is able to learn the concatenated $F_{avg}^{s}$ and $F_{max}^{s}$ to generate a spatial attention map. In this work, the spatial attention module ($\operatorname{SAM}$) can be given as follows
    \begin{equation}
	\begin{aligned}
		\operatorname{SAM} & =\sigma\left(\mathrm{Conv}^{7 \times 7}([\mathrm{Avg}(x_{in}^s) ; \mathrm{Max}(x_{in}^s)])\right) \\
		& =\sigma\left(\mathrm{Conv}^{7 \times 7}\left(\left[F_{avg}^{s} ; F_{max}^{s}\right]\right)\right),
	\end{aligned}
    \end{equation}
    where $x_{in}^s$ is the input of $\operatorname{SAM}$, $\left[~;~\right]$ is exploited to concatenate two types of pooled features.

    The joint application of channel attention and spatial attention can learn features more comprehensively and improve the model performance and interpretability. Moreover, the unnecessary calculations can be diminished, increasing the operational efficiency, since the proposed method mainly focuses on the critical channels and regions. The ablation experiments in Section \ref{as} will verify the importance of these two types of attention mechanisms in our learning network.
\subsection{Structural Reparameterization}
\begin{figure}[t]
        \centering
        \includegraphics[width=1.00\linewidth]{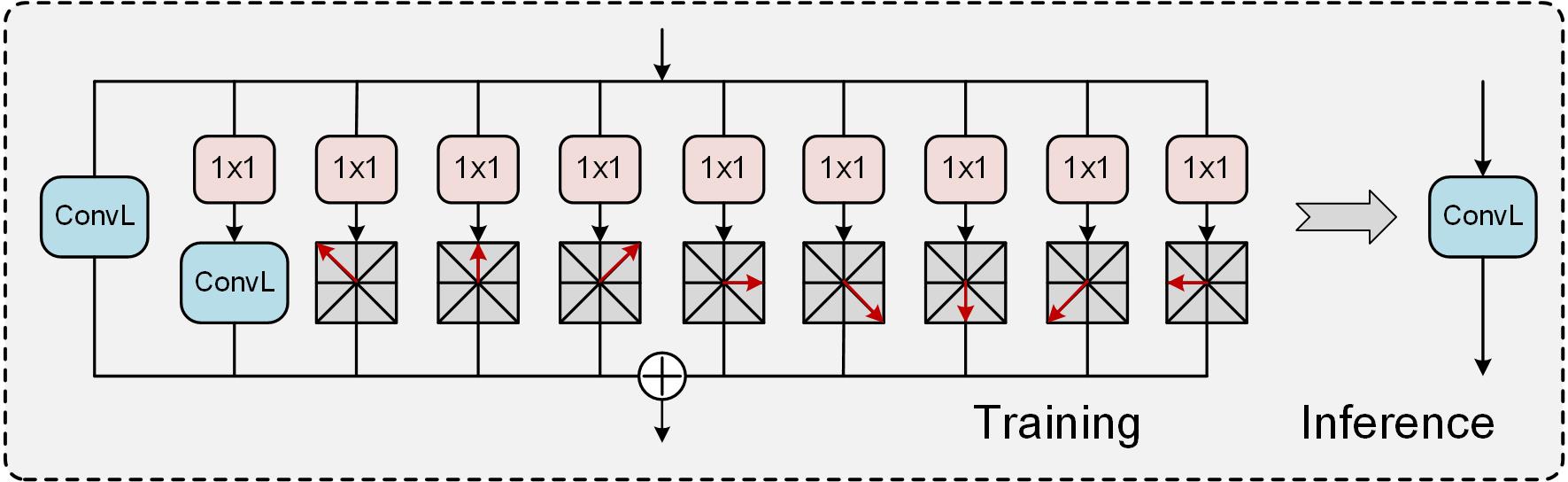}
        \caption{The pipeline of Kirsch-guided reparameterization module (KRM). In the training stage, the KRM employs multiple branches, which can be merged into one normal convolutional layer in the inference stage.}
        \label{Figure03}
    \end{figure}

    To efficiently deploy the enhanced method on edge devices such as ships and maritime video surveillance systems, the deep network needs to be lightweight, thereby reducing the computation and improving the processing speed of a single image frame. Edge detection operators have been successfully applied in image denoising, super-resolution reconstruction, low-light image enhancement, and other fields. Zhang \textit{et al.} \cite{zhang2021edge} propose to combine the Sobel and Laplacian filters into deep neural networks. However, considering the complexity and variability of rain streaks and potential edge features, it is difficult to accurately eliminate unwanted edge features with gradient information in vertical and horizontal directions. The reparameterization has achieved satisfactory results in different vision tasks. To eliminate rain streaks in different directions and extract complex and variable edge texture structures, this paper designs a more suitable Kirsch-guided reparameterization module (KRM) with shared parameters. As shown in Fig. \ref{Figure03}, KRM mainly consists of a normal convolutional layer, an expanding-and-squeezing convolutional layer, and the edge detection operators in eight directions which learn and infer network parameters. The normal convolutional and expanding-and-squeezing convolutional operations can be given as follows
    \begin{equation}\label{eq:F_n}
	\begin{gathered}
		F_n=W_n * x_{i n}^k+B_n,
	\end{gathered}
   \end{equation}
   \begin{equation}\label{eq:F_es}
	\begin{gathered}
            F_{e s}=W_s *\left(W_e * x_{i n}^k+B_e\right)+B_s,
	\end{gathered}
   \end{equation}
   where $x_{i n}^k$ represents the input of KRM. $W_n$, $W_s$, $W_e$, $B_n$, $B_s$, $B_e$ are the weights and bias of corresponding convolution. $F_n$ is the output of the normal convolutional layer. $F_{es}$ is the output of the expanding-and-squeezing convolutional layer. The subscripts $n$, $e$, and $s$ represent the normal, expanding, and squeezing items, respectively. We first reparameterize the Eqs. (\ref{eq:F_n}) and (\ref{eq:F_es}), and then merge them into one single normal convolution with parameters $W_{es}$ and $B_{es}$, i.e.,
	\begin{equation}
		\begin{aligned}
			& W_{e s} = \mathrm{perm} \left(W_e\right) * W_s,
		\end{aligned}
	\end{equation}
	\begin{equation}
		\begin{aligned}
			& B_{e s}=W_s * \mathrm{rep} \left(B_e\right)+B_s,
		\end{aligned}
	\end{equation}
 where $\operatorname{perm}(\cdot)$ denotes the permute operation which exchanges the $1$st and $2$nd dimensions of a tensor, $\operatorname{rep} ( \cdot )$ denotes the spatial broadcasting operation, which replicates the bias $B_e \in \mathbb{R}^{1 \times {D} \times 1 \times 1}$ into $\operatorname{rep}\left(B_e\right) \in \mathbb{R}^{1 \times D \times 3 \times 3}$.

    We propose to incorporate the predefined eight-direction Kirsch edge filters $K_i$, shown in Table \ref{Table_Kirsch}, into the reparameterization module. To memorize the edge features, the input feature $x_{in}^{k}$ will first be processed by $C\times C\times 1\times 1$ convolution and then use a custom Kirsch filter to extract the feature map gradients in eight different directions. To improve the correlation of features between different channels and reduce the amount of computation, we set a scaling factor, which is set to $2$ in our experiments, to scale the channels during the learning process. Therefore, the edge information in eight directions can be expressed as follows
	\begin{equation}
		\begin{aligned}
			F_{K}^i&=\left(S_{K}^{i} \odot K_i\right) \otimes \left( W_i * x_{in}^{k} + B_i \right)+B_{K_i}\\
			& =W_{K}^i \otimes\left(W_i * x_{in}^{k}+B_i\right)+B_{K_i},
		\end{aligned}        
	\end{equation}
 where $W_i$ and $B_i$ are the weights and bias of $1 \times 1$ convolution for branches in eight directions, $S_{K}^{i}$ and $B_{K_i}$ are the scaling parameters and bias with the shape of $C\times 1 \times 1 \times 1$, $\odot$ indicates the channel-wise broadcasting multiplication, $ S_{K}^{i} \odot K_i $ is formed in the shape of $C \times 1 \times 3 \times 3$, $\otimes$ and $*$, respectively, represent the depth-wise convolution and normal convolution. The combined edge information, extracted by the scaled Kirsch filters, is given by
	\begin{equation}
		F_{\mathrm{K}}= \sum_{i=1}^{8} F_{K}^i.
	\end{equation}

	Therefore, the final weights and bias after reparameterization can be expressed as follows
	\begin{equation}
		\begin{aligned}
			W_{\mathrm{rep}} & =W_n+W_{es}+\sum_{i=1}^{8} \left( \mathrm{perm} \left(W_i\right) * W_{\mathrm{K}}^i \right),
		\end{aligned}
	\end{equation}
	\begin{equation}
		\begin{aligned}
			B_{\mathrm{rep}} & =B_n+B_{e s}+\sum_{i=1}^{8} \left( \mathrm{perm} \left(B_i\right) * B_{\mathrm{K}_i} \right).
		\end{aligned}
	\end{equation}
	The output feature can be obtained using one single normal convolution in the inference stage, i.e.,
	\begin{equation}
		F=W_{\mathrm{rep }} * x_{in}^{k} + B_{\mathrm{rep}}.
	\end{equation}
	By reparameterizing the gradient features extracted by Kirsch operators into a single convolutional layer, the proper balance between edge extraction and computational time can be achieved in numerical experiments.
 \subsection{Basic Learning Block}
   We propose to construct an edge-driven attention residual block (termed EARB), consisting mainly of convolutional layers, channel attention, and spatial attention. As shown in Fig. \ref{Figure02}, the Kirsch-guide reparameterization module could provide meaningful gradient information, thus making EARBs more sensitive to edge information. It can effectively extract valuable edge features from original images. In this work, we only exploit five EARBs to build the low-visibility enhancement network. Our ERANet achieves satisfactory restoration in three common low-visibility scenarios with only $2.4$MB of network parameters. In addition, benefiting from the lightweight design, the ERANet only takes $0.016$ seconds to process a single image with a resolution of $1920 \times 1080$ pixels (i.e., $1080$p).
	\setlength{\tabcolsep}{4.25pt}
	\begin{table}[t]
		\centering
		\caption{The details of training and testing datasets used in our experiments.}
		\begin{tabular}{l|cc|ccc}
			\hline
 			Datasets                             & Train & Test & Dehazing         & Deraining         & \tabincell{c}{Low-Light \\ Enhancement} \\ \hline \hline
			RESIDE-OTS \cite{li2018benchmarking} & 1500  & 50   & \CheckmarkBold &                &                       \\
			Rain100L \cite{yang2019joint}        & 200   & 100  &                & \CheckmarkBold &                       \\
			LOL \cite{wei2018deep}               & 1485  & 15   &                &                & \CheckmarkBold        \\
			Seaships \cite{shao2018seaships}     & 1000  & 300  & \CheckmarkBold & \CheckmarkBold & \CheckmarkBold        \\ 
			SMD \cite{prasad2017video}           & 1000  & 300  & \CheckmarkBold & \CheckmarkBold & \CheckmarkBold        \\
			\hline
		\end{tabular}
		\label{Table_datasets}
	\end{table}
	\setlength{\tabcolsep}{3.8pt}
	\begin{table}[t]
		\centering
		\caption{Methods for comparison with ERANet.}
		\begin{tabular}{l|cccc}
			\hline
			Methods                                        & Publication  & Dehazing       & Deraining      & \tabincell{c}{Low-Light \\ Enhancement} \\ \hline\hline
			DCP \cite{he2010single}                        & TPAMI (2010) & \CheckmarkBold &                &                 \\ 
			NPE \cite{wang2013naturalness}                 & TIP (2013)   &                &                & \CheckmarkBold \\ 
			SDD \cite{hao2020low}                          & TMM (2020)   & \CheckmarkBold &                & \CheckmarkBold \\ 
			ROP+ \cite{liu2022rank}                        & TPAMI (2023) & \CheckmarkBold &                & \CheckmarkBold \\ \hline
			DDN \cite{fu2017removing}                      & CVPR (2017)  &                & \CheckmarkBold &                \\ 
			RetinexNet \cite{wei2018deep}                  & BMVC (2018)  &                &                & \CheckmarkBold \\ 
			KinD \cite{zhang2019kindling}                  & MM (2019)    &                &                & \CheckmarkBold \\ 
			LPNet \cite{fu2019lightweight}                 & TNNLS (2019) &                & \CheckmarkBold &                \\ 
			GCANet \cite{chen2019gated}                    & WACV (2019)  & \CheckmarkBold & \CheckmarkBold &                \\ 
			DIG \cite{ran2020single}                       & ICME (2020)  &                & \CheckmarkBold &                \\      
			DualGCN \cite{fu2021rain}                      & AAAI (2021)  &                & \CheckmarkBold &                \\ 
			LLFlow \cite{wang2022low}                      & AAAI (2022)  &                &                & \CheckmarkBold \\ 
			TSDNet \cite{liu2022deep}                      & TII (2022)   & \CheckmarkBold &                &                \\ 
			AirNet \cite{li2022all}                        & CVPR (2022)  & \CheckmarkBold & \CheckmarkBold &                \\ 
			MIRNetv2 \cite{zamir2022learning}              & TPAMI (2022) & \CheckmarkBold & \CheckmarkBold & \CheckmarkBold \\ 
			TransWeather \cite{valanarasu2022transweather} & CVPR (2022)  & \CheckmarkBold & \CheckmarkBold & \CheckmarkBold \\ 
			SMNet \cite{lin2023smnet}                      & TMM (2023)   &                &                & \CheckmarkBold \\                         
			KBNet \cite{zhang2023kbnet}                    & Arxiv (2023) &                & \CheckmarkBold &                \\ 
			USCFormer \cite{wang2023uscformer}             & TITS (2023)  & \CheckmarkBold &                &                \\ 
			WeatherDiff \cite{ozdenizci2023restoring}      & TPAMI (2023) & \CheckmarkBold & \CheckmarkBold & \CheckmarkBold \\ \hline
			ERANet                                         &  ---         & \CheckmarkBold & \CheckmarkBold & \CheckmarkBold \\ \hline
		\end{tabular}
		\label{C-Methods}
	\end{table}
    \subsection{Loss Function}\label{loss}
    To meet the requirements of three different low-visibility scene restoration tasks, we propose to develop a hybrid loss function to preserve meaningful information, such as color and textural features, etc. It mainly includes three parts, i.e., multi-scale structural similarity loss $\mathcal{L}_{\emph{\text{MS-SSIM}}}$, $\ell_1$-norm loss $\mathcal{L}_{\ell_1}$, and total variation loss $\mathcal{L}_{TV}$, i.e.,
    \begin{equation}
		\mathcal{L}_{\text{total}}= \gamma_{1}\cdot\mathcal{L}_{\emph{\text{MS-SSIM}}} + 
		\gamma_{2}\cdot\mathcal{L}_{\ell_1} + \gamma_{3}\cdot\mathcal{L}_{TV},
		\label{eq:loss}
    \end{equation}
    where $\gamma_{1}$, $\gamma_{2}$, and $\gamma_{3}$ are positive weights. According to the extensive experiments, we empirically select the parameters $\gamma_{1} = 0.85$, $\gamma_{2} = 0.15$, and $\gamma_{3} = 0.01$ in this work. Comprehensive experiments on multi-scene visibility enhancement have demonstrated the robustness and effectiveness of these manually-selected parameters.
    \subsubsection{Multi-Scale Structural Similarity Loss}
    The multi-scale structural similarity (MS-SSIM) derives from the original structural similarity (SSIM) at different scales. To be specific, the SSIM is defined as follows
    \begin{equation}
        \begin{aligned}
        \operatorname{SSIM}(\hat{H}, H) & =\frac{2 \mu_{\hat{H}} \mu_H+c_1}{\mu_{\hat{H}}^2+\mu_H^2+c_1} \cdot \frac{2 \sigma_{\hat{H}H}+c_2}{\sigma_{\hat{H}}^2+\sigma_H^2+c_2} \\
        & =l(x) \cdot {cs}(x),
        \end{aligned}
    \end{equation}
    where $x$ denotes the pixel index, $\hat{H}$ represents the output of network, $H$ represents the ground truth, $\mu_{\hat{H}}$ and $\mu_H$ represent the local averages, $\sigma_{\hat{H}}$ and $\sigma_H$ mean the standard deviations, $\sigma_{\hat{H}H}$ represents the covariance value, $c_1$ and $c_2$ are constant parameters to avoid instability.

    Finally, we define $\mathcal{L}_{\emph{\text{MS-SSIM}}}$, one part of the total loss function of ERANet, as follows
    \begin{equation}
    \mathcal{L}_{\emph{\text{MS-SSIM}}}=1-l_\mathcal{M} \cdot \prod_{j=1}^\mathcal{M}\left[c s_j\right]^{\beta_j},
    \end{equation}
    where $\mathcal{M}$ denotes the default parameter of scales. Please refer to \cite{wang2003multiscale} for more details about MS-SSIM.
    \subsubsection{$\ell_1$-norm Loss}
    To guarantee the imaging quality, we propose to adopt the $\ell_1$-norm as one part of our loss function in ERANet, which is given by
    \begin{equation}
          \mathcal{L}_{\ell_1} = \left\| \hat{H} - H \right\|_1.
    \end{equation}
    \subsubsection{Total Variation Loss}
    The total variation (TV) loss (i.e., $\mathcal{L}_{TV}$) is also suggested to enhance the spatial smoothness of the generated image. It will not affect the high-frequency part of the image due to its small weight occupancy. It promotes the smoothness of the reconstructed image by penalizing sudden changes in pixel values, resulting in visually pleasing and artifact-free results. Therefore, $\mathcal{L}_{TV}$ can be defined as
    \begin{equation}
        \mathcal{L}_{TV} = \left\|\nabla_h \hat{H} \right\|_2 + \left\| \nabla_w \hat{H} \right\|_2,
    \end{equation}
     where $\nabla_h$ and $\nabla_w$ are operators to compute the horizontal and vertical gradients of $\hat{H}$. The weight of $\mathcal{L}_{TV}$ is only set to $0.01$ in our numerical experiments, as a larger weight would cause the overly-smoothing effects in the restored images.   
    \begin{figure}[t]
        \centering
        \includegraphics[width=1.00\linewidth]{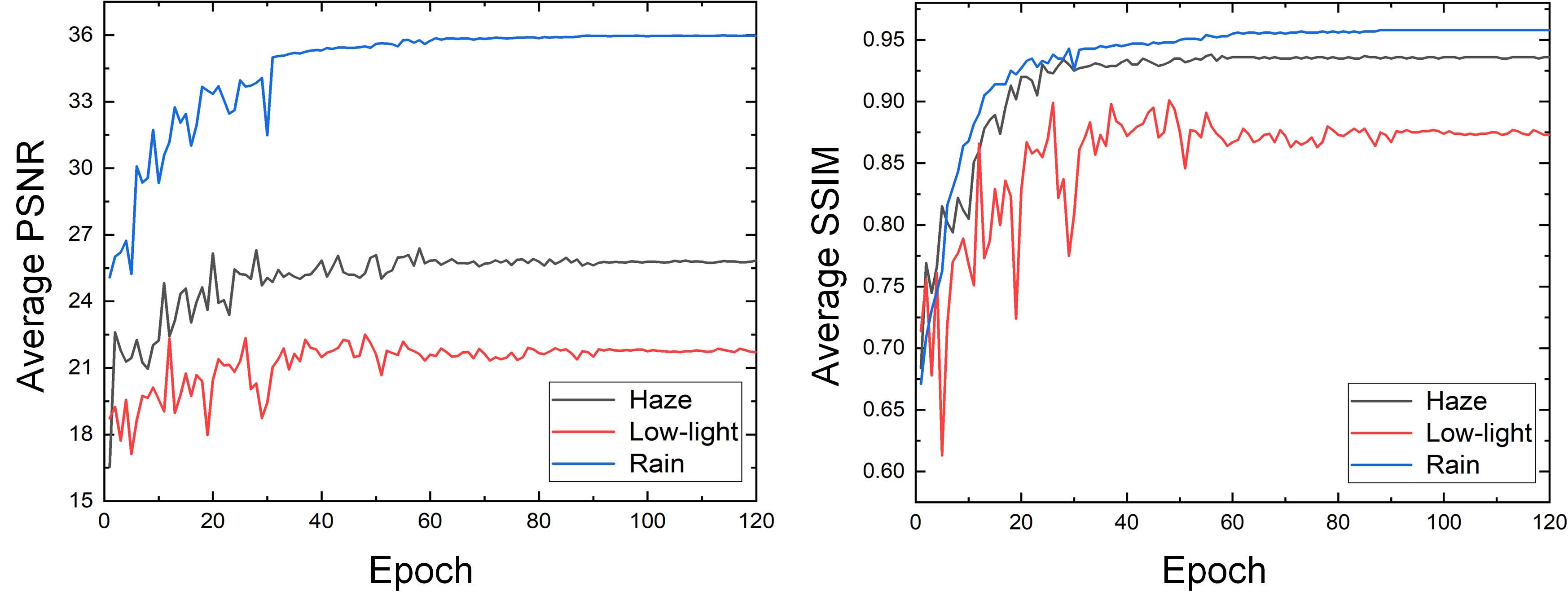}
        \caption{The convergence analysis under different degradation scenarios.}
        \label{Figure_epoch}
    \end{figure}
\section{Experiments and Discussion}\label{exper}
In this section, we first introduce the experimental implementation details, which include train/test datasets, evaluation metrics, competitive methods, and the experimental platform. To demonstrate the superiority of ERANet, both quantitative and qualitative comparisons with state-of-the-art methods on standard and IWTS-related datasets are presented. To validate the network's rationality, we have conducted several ablation experiments. The experiments on YOLOv7-based \cite{wang2022yolov7} vessel detection, DeepLabv3+-based \cite{chen2018encoder} scene segmentation, model size, and running time are performed to demonstrate that our real-time scene recovery method for reconstructing low-visibility images has huge potential for promoting the navigational safety of vessels in IWTS. Our source code is freely available at \url{https://github.com/LouisYuxuLu/ERANet}.
    \setlength{\tabcolsep}{7.0pt}	
    \begin{table}[t]
    	\centering
    	\caption{PSNR, SSIM, and NIQE results of various dehazing methods on RESIDE-OTS \cite{li2018benchmarking}, Seaships \cite{shao2018seaships}, and SMD \cite{prasad2017video}. The best three results are highlighted in {\color{red}red}, {\color{blue}blue}, and {\color{green}green} colors, respectively.}
    	\begin{tabular}{l|ccc}
    		\hline
    		Methods                                        & PSNR $\uparrow$             & SSIM $\uparrow$              & NIQE $\downarrow$ \\\hline\hline
    		DCP \cite{he2010single}                        & 15.128$\pm$3.607 & 0.823$\pm$0.075 & 5.162$\pm$1.962 \\ 
    		SDD \cite{hao2020low}                          & 14.831$\pm$2.746 & 0.842$\pm$0.086 & 6.384$\pm$1.901 \\ 
    		ROP+ \cite{liu2022rank}                        & 17.474$\pm$4.335 & 0.883$\pm$0.067 & 5.504$\pm$1.916 \\ 
    		GCANet \cite{chen2019gated}                    & 17.210$\pm$4.454 & 0.885$\pm$0.056 & \color{red}{4.611$\pm$0.722} \\ 
    		TSDNet \cite{liu2022deep}                      & 19.061$\pm$3.635 & \color{green}{0.908$\pm$0.064} & 5.115$\pm$1.401 \\ 
    		AirNet \cite{li2022all}                        & 15.476$\pm$3.901 & 0.663$\pm$0.127 & 5.016$\pm$1.111 \\
    		MIRNetv2 \cite{zamir2022learning}              & 20.484$\pm$6.231 & \color{blue}{0.909$\pm$0.075} & \color{green}{4.909$\pm$1.140} \\ 
    		TransWeather \cite{valanarasu2022transweather} & \color{blue}{21.922$\pm$5.481} & 0.900$\pm$0.069 & 5.388$\pm$1.377 \\
    		USCFormer \cite{wang2023uscformer}             & \color{green}{21.533$\pm$4.373} & 0.889$\pm$0.100 & 5.622$\pm$1.661 \\
    		WeatherDiff \cite{ozdenizci2023restoring}      & 16.848$\pm$2.851 & 0.890$\pm$0.066 & 5.435$\pm$1.297 \\ \hline
    		ERANet                                         & \color{red}{24.595$\pm$5.134}   & \color{red}{0.946$\pm$0.046} & \color{blue}{4.718$\pm$1.212} \\ \hline
    	\end{tabular}\label{Table_Haze}
    \end{table}
    \setlength{\tabcolsep}{7.0pt}	
    \begin{table}[t]
    	\centering
    	\caption{PSNR, SSIM, and NIQE results of various deraining methods on Rain100L \cite{yang2019joint}, Seaships \cite{shao2018seaships}, and SMD \cite{prasad2017video}. The best three results are highlighted in {\color{red}red}, {\color{blue}blue}, and {\color{green}green} colors, respectively.}
    	\begin{tabular}{l|c|c|c}
    		\hline
    		Methods                                        & PSNR $\uparrow$             & SSIM $\uparrow$              & NIQE $\downarrow$ \\\hline\hline
    		DDN \cite{fu2017removing}                      & 28.934$\pm$2.943 & 0.908$\pm$0.044 & 5.211$\pm$1.110 \\ 
    		LPNet \cite{fu2019lightweight}                 & 31.980$\pm$2.719 & 0.946$\pm$0.022 & \color{blue}{4.816$\pm$1.340} \\ 
    		DIG \cite{ran2020single}                       & 31.621$\pm$2.533 & 0.936$\pm$0.023 & 5.047$\pm$1.213 \\ 
    		GCANet \cite{chen2019gated}                    & 16.387$\pm$5.593 & 0.702$\pm$0.103 & 5.194$\pm$1.006 \\ 
    		DualGCN \cite{fu2021rain}                      & \color{blue}{36.072$\pm$2.763} & \color{red}{0.969$\pm$0.014} & 5.406$\pm$1.559 \\ 
    		AirNet \cite{li2022all}                        & 29.618$\pm$5.711 & 0.892$\pm$0.080 & \color{green}{4.989$\pm$1.248} \\ 
    		MIRNetv2 \cite{zamir2022learning}              & 26.732$\pm$3.717 & 0.866$\pm$0.065 & 5.189$\pm$1.164 \\ 
    		TransWeather \cite{valanarasu2022transweather} & 24.821$\pm$3.196 & 0.870$\pm$0.060 & 5.278$\pm$1.221 \\ 
    		KBNet \cite{zhang2023kbnet}                    & \color{red}{36.304$\pm$3.608} & \color{blue}{0.962$\pm$0.023} & 5.973$\pm$1.821 \\ 
    		WeatherDiff \cite{ozdenizci2023restoring}      & 20.677$\pm$2.009 & 0.879$\pm$0.062 & 5.150$\pm$1.176 \\ \hline
    		ERANet                                         & \color{green}{34.436$\pm$3.765} & \color{green}{0.962$\pm$0.027} & \color{red}{4.687$\pm$1.283} \\ \hline
    	\end{tabular}\label{Table_Rain}
    \end{table}
    \setlength{\tabcolsep}{7.0pt}	
    \begin{table}[t]
    	\centering
    	\caption{PSNR, SSIM, and NIQE results of various low-light enhancement methods on LOL \cite{wei2018deep}, Seaships \cite{shao2018seaships}, and SMD \cite{prasad2017video}. The best three results are highlighted in {\color{red}red}, {\color{blue}blue}, and {\color{green}green} colors, respectively.}
    	\begin{tabular}{l|c|c|c}
    		\hline
    		Methods                                        & PSNR $\uparrow$             & SSIM $\uparrow$              & NIQE $\downarrow$ \\\hline\hline
    		NPE \cite{wang2013naturalness}                 & 14.647$\pm$4.870 & 0.726$\pm$0.169 & \color{red}{4.880$\pm$1.464} \\ 
    		SDD \cite{hao2020low}                          & 14.874$\pm$4.477 & 0.718$\pm$0.176 & 5.707$\pm$1.350 \\ 
    		ROP+ \cite{liu2022rank}                        & 11.955$\pm$2.935 & 0.590$\pm$0.212 & 5.286$\pm$1.380 \\ 
    		RetinexNet \cite{wei2018deep}                  & \color{blue}{16.560$\pm$2.325} & \color{blue}{0.820$\pm$0.080} & 5.388$\pm$1.260 \\ 
    		KinD \cite{zhang2019kindling}                  & \color{green}{16.397$\pm$4.514} & \color{green}{0.772$\pm$0.169} & 5.277$\pm$1.365 \\ 
    		LLFlow \cite{wang2022low}                      & 13.188$\pm$3.782 & 0.719$\pm$0.110 & 5.616$\pm$1.194 \\ 
    		MIRNetv2 \cite{zamir2022learning}              & 12.301$\pm$3.854 & 0.624$\pm$0.191 & \color{green}{5.165$\pm$0.880} \\ 
    		TransWeather \cite{valanarasu2022transweather} & 13.187$\pm$3.889 & 0.699$\pm$0.136 & 5.481$\pm$1.067 \\ 
    		SMNet \cite{lin2023smnet}                      & 14.790$\pm$5.154 & 0.728$\pm$0.163 & 5.339$\pm$1.101 \\ 
    		WeatherDiff \cite{ozdenizci2023restoring}      & 12.916$\pm$2.610 & 0.727$\pm$0.120 & 5.257$\pm$0.753 \\ \hline
    		ERANet                                         & \color{red}{20.877$\pm$5.103} & \color{red}{0.917$\pm$0.067} & \color{blue}{4.902$\pm$1.182} \\ \hline
    	\end{tabular}\label{Table_Low}
    \end{table}
\subsection{Implementation Details}
\subsubsection{Datasets}
   The pairs (i.e., clear and low-visibility) of real-world IWTS-related images are difficult to obtain in maritime scenarios. It inevitably brings great challenges to the learning-based imaging networks. We thus apply the IWTS-related dataset Seaships \cite{shao2018seaships} and Singapore Maritime Dataset (SMD) \cite{prasad2017video} to synthesize low-visibility images through Eqs. (\ref{Equation_atmospheric scattering model})-(\ref{Equation_low-light model}). To verify the robustness and generalization ability of our method, we also conduct experiments on standard datasets, which include RESIDE-OTS \cite{li2018benchmarking} (dehazing), Rain100L \cite{yang2019joint} (deraining), and LOL \cite{wei2018deep} (low-light enhancement). The specific information of the datasets used to train and test our ERANet is shown in Table \ref{Table_datasets}.
\subsubsection{Evaluation Metrics}
   To quantitatively evaluate the visibility enhancement results, we employ the peak signal-to-noise ratio (PSNR) and structural similarity (SSIM) \cite{wang2003multiscale} as the reference-based evaluation metrics. We also apply the no-reference natural image quality evaluator (NIQE) \cite{NIQE} to assess the objective performance of our and other competitive methods. It is worth noting that larger PSNR and SSIM values and smaller NIQE values represent better scene recovery.
\subsubsection{Competitive Methods}
   To assess the performance of low-visibility scene recovery, shown in Table \ref{C-Methods}, we compare ERANet with several state-of-the-art imaging methods. To evaluate the scene versatility of ERANet, we select several advanced methods (such as ROP+ \cite{liu2022rank} and TransWeather \cite{valanarasu2022transweather}) that can restore two or three types of low-visibility scenes. For the impartiality and fairness, all competitive methods are derived from the source codes released by the authors.
\subsubsection{Experimental Platform}\label{sssec:ep}
   We train the network for $120$ epochs using the Adam optimizer. The initial learning rate of the optimizer is $0.001$, which is multiplied by $0.1$ after every $30$ epochs. As shown in Fig. \ref{Figure_epoch}, we conduct network convergence analysis on the standard datasets related to three scenarios (i.e., RESIDE-OTS \cite{li2018benchmarking}, Rain100L \cite{yang2019joint}, LOL \cite{wei2018deep}). We can observe that the network is converged with $90$ epochs, and the subsequent training process demonstrates stable network performance. The learning network is trained and tested in a Python 3.7 environment using the PyTorch software package with a PC with Intel(R) Core(TM) i9-12900K CPU @2.30GHz and Nvidia GeForce RTX 3080 Ti Laptop GPU.
   \begin{figure*}[t]
	\centering
        \includegraphics[width=1.00\linewidth]{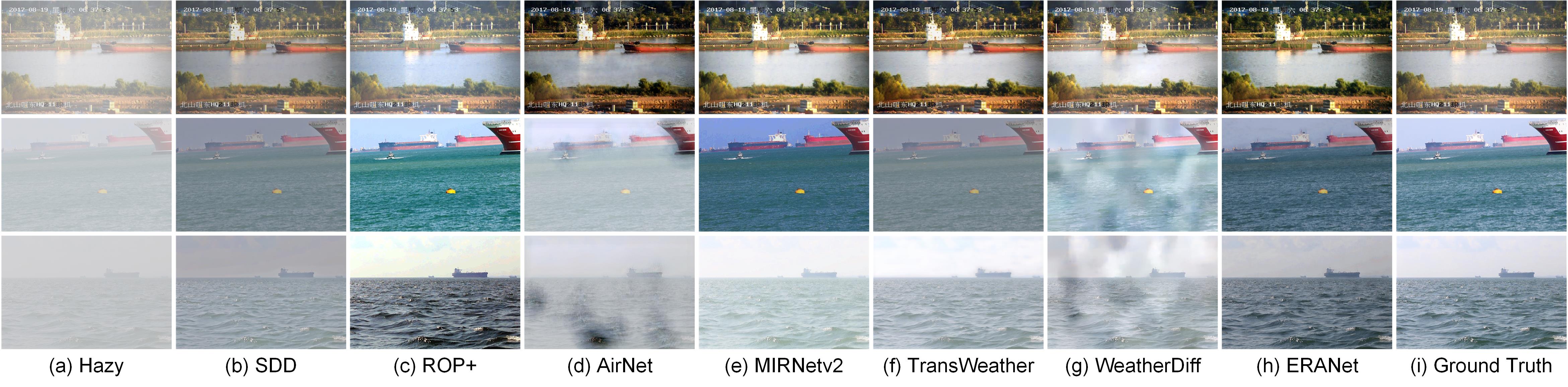}
        \caption{Visual comparisons of dehazing results from Seaships \cite{shao2018seaships} and SMD \cite{prasad2017video}. From left to right: (a) hazy images, restored images, respectively, yielded by (b) SDD \cite{hao2020low}, (c) ROP+ \cite{liu2022rank}, (d) AirNet \cite{li2022all}, (e) MIRNetv2 \cite{zamir2022learning}, (f) TransWeather \cite{valanarasu2022transweather}, (g) WeatherDiff \cite{ozdenizci2023restoring}, (h) ERANet, and (i) Ground Truth.}
        \label{Figure_haze}
    \end{figure*}
	\begin{figure*}[t]
		\centering
		\includegraphics[width=1.00\linewidth]{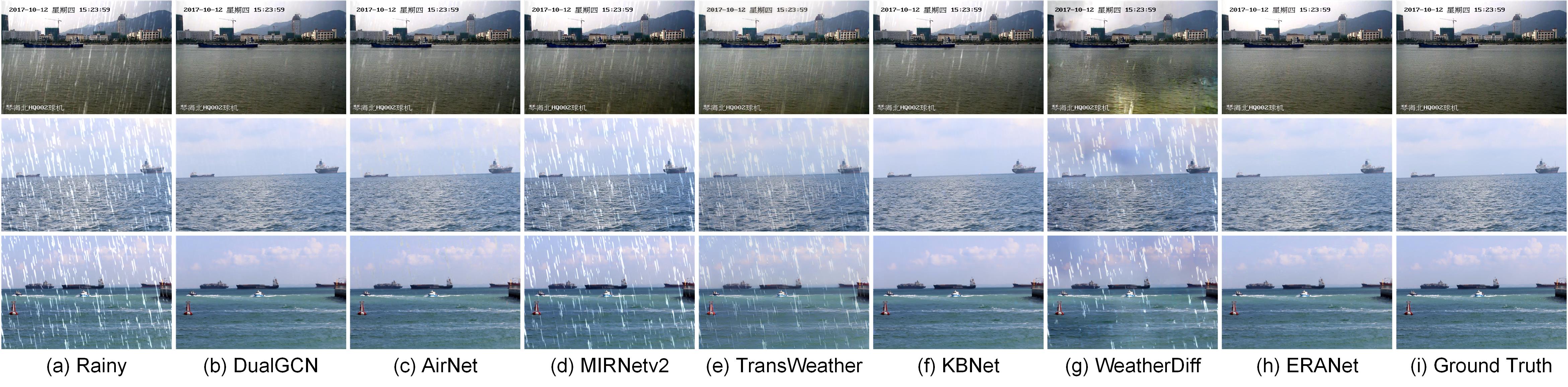}
		\caption{Visual comparisons of deraining results from Seaships \cite{shao2018seaships} and SMD \cite{prasad2017video}. From left to right: (a) rainy images, restored images, respectively, yielded by (b) DualGCN \cite{fu2021rain}, (c) AirNet \cite{li2022all}, (d) MIRNetv2 \cite{zamir2022learning}, (e) TransWeather \cite{valanarasu2022transweather}, (f) KBNet \cite{zhang2023kbnet}, (g) WeatherDiff \cite{ozdenizci2023restoring}, (h) ERANet, and (i) Ground Truth.}
		\label{Figure_rain}
	\end{figure*}
	\begin{figure*}[t]
		\centering
		\includegraphics[width=1.00\linewidth]{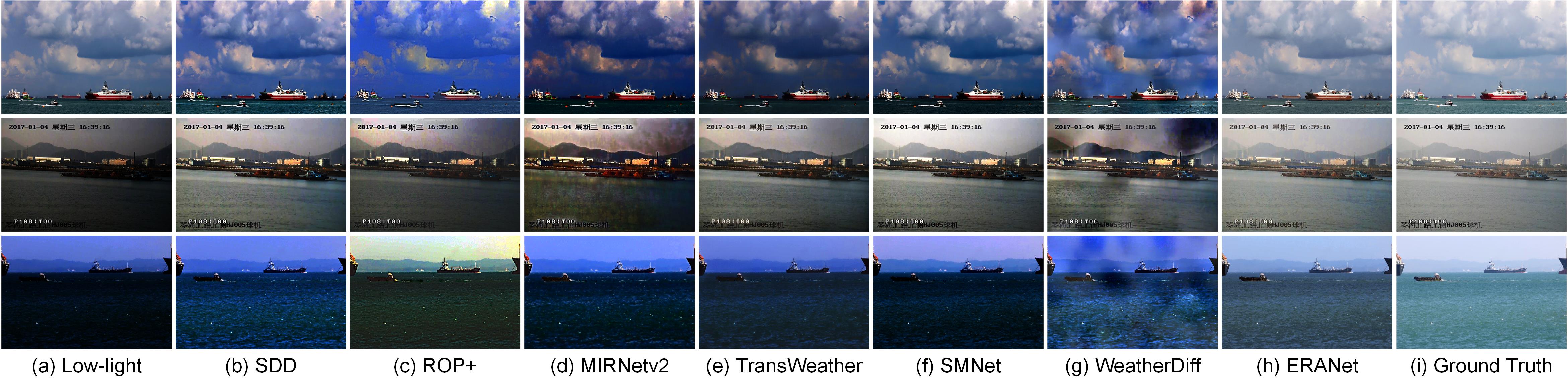}
		\caption{Visual comparisons of low-light enhancement results from Seaships \cite{shao2018seaships} and SMD \cite{prasad2017video}. From left to right: (a) Low-light images, restored images, respectively, yielded by (b) SDD \cite{hao2020low}, (c) ROP+ \cite{liu2022rank},  (d) MIRNetv2 \cite{zamir2022learning}, (e) TransWeather \cite{valanarasu2022transweather}, (f) SMNet \cite{lin2023smnet}, (g) WeatherDiff \cite{ozdenizci2023restoring}, (h) ERANet, and (i) Ground Truth.}
		\label{Figure_low}
	\end{figure*}
	\begin{figure*}[t]
		\centering
		\includegraphics[width=1.00\linewidth]{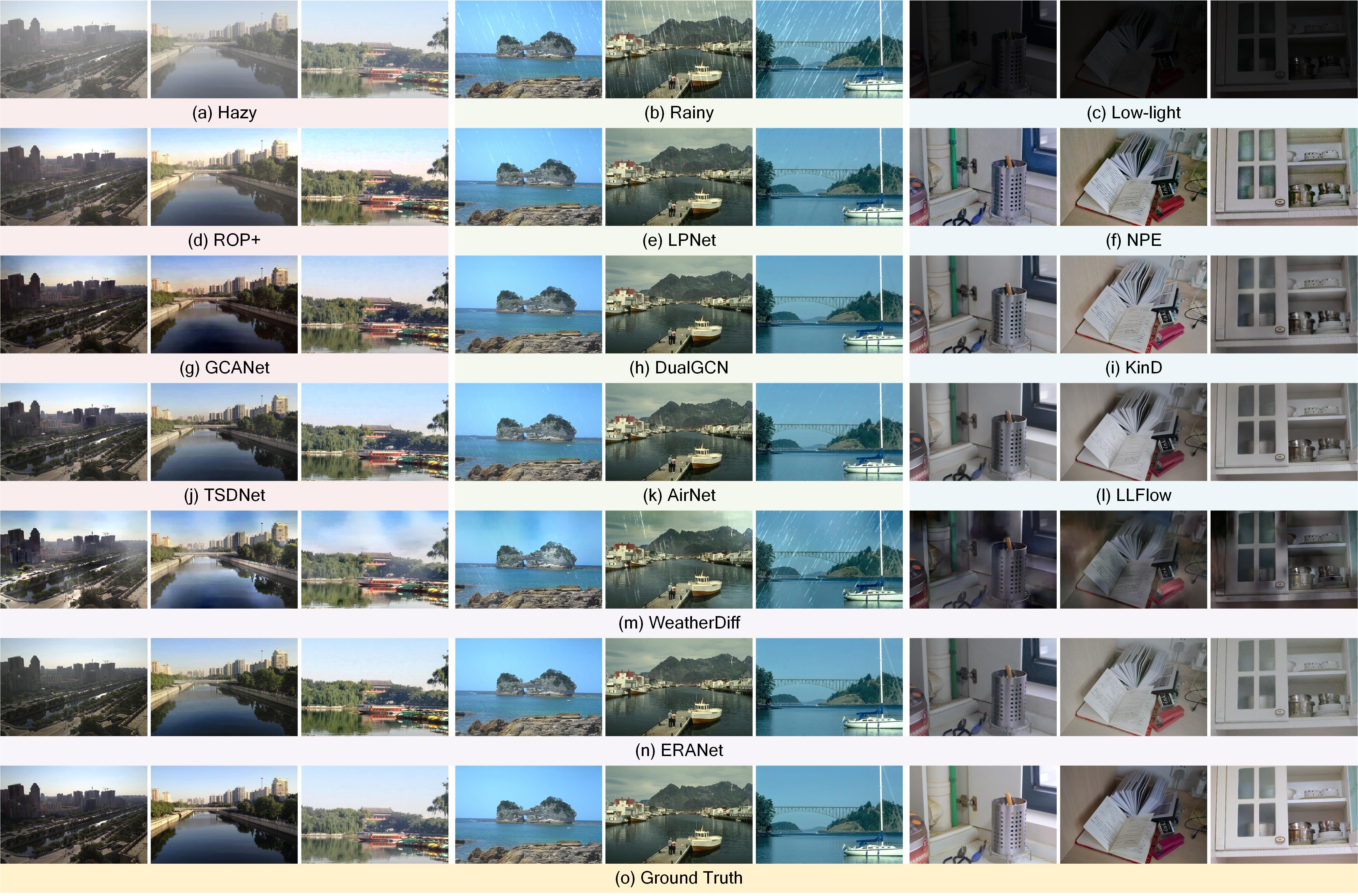}
		\caption{Visual comparisons of multi-scene recovery performance from three standard test datasets (i.e., RESIDE-OTS \cite{li2018benchmarking}, Rain100L \cite{yang2019joint}, and LOL \cite{wei2018deep}). From top to bottom: (a) hazy images, (b) rainy images, (c) low-light images, restored images, respectively, generated by (d) ROP+ \cite{liu2022rank}, (e) LPNet \cite{fu2019lightweight}, (f) NPE \cite{wang2013naturalness}, (g) GCANet \cite{chen2019gated}, (h) DualGCN \cite{fu2021rain}, (i) KinD \cite{zhang2019kindling}, (j) TSDNet \cite{liu2022deep}, (k) AirNet \cite{li2022all}, (l) LLFlow \cite{wang2022low}, (m) WeatherDiff \cite{ozdenizci2023restoring}, (n) ERANet, and (o) Ground Truth.}
		\label{Figure07_SD}
	\end{figure*}
        \begin{figure*}[t]
            \centering
		\includegraphics[width=1.00\linewidth]{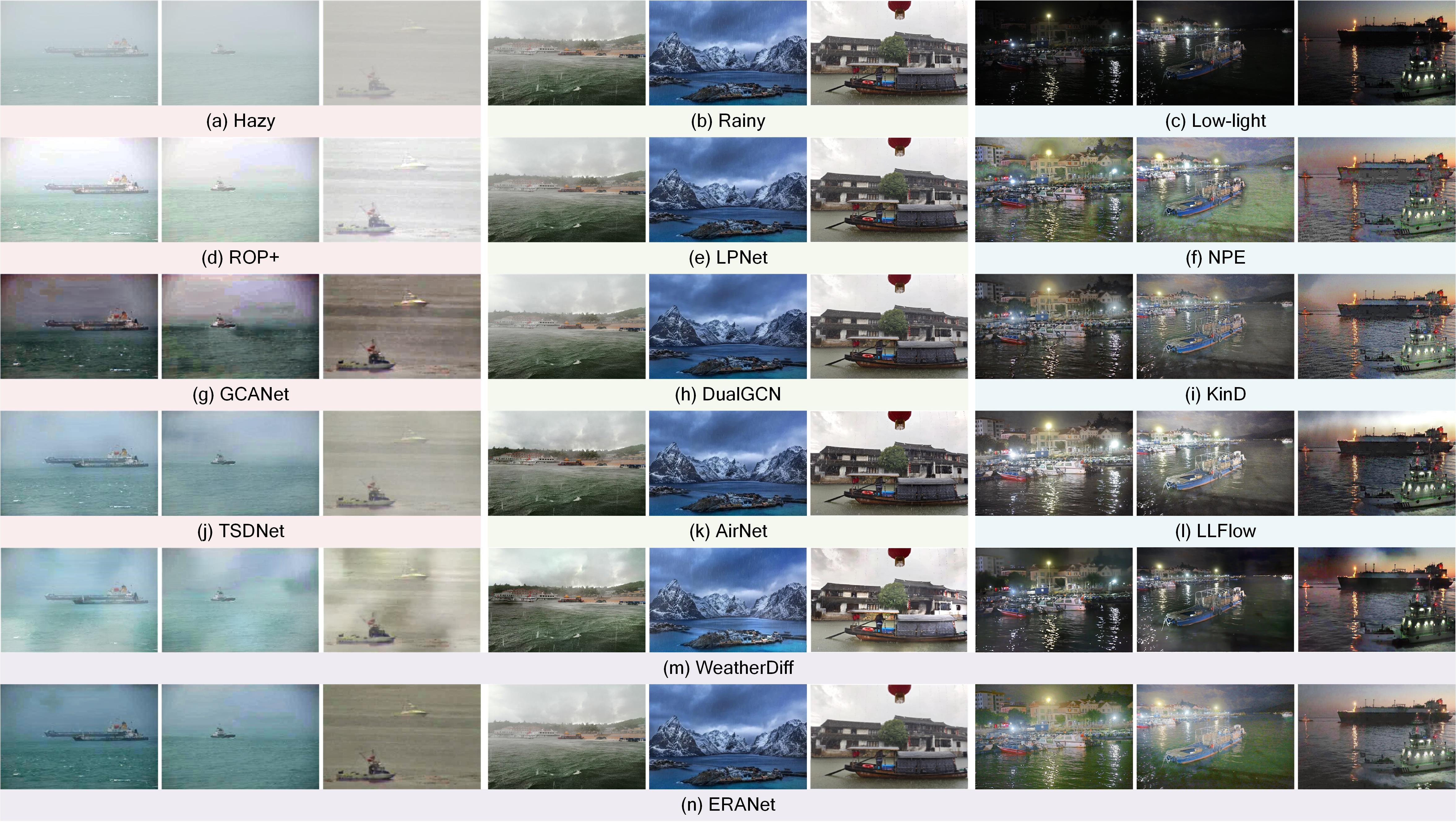}
		\caption{Visual comparisons of real-world scene recovery performance from three types of low-visibility images. From top to bottom: (a) hazy images, (b) rainy images, (c) low-light images, restored images, respectively, generated by (d) ROP+ \cite{liu2022rank}, (e) LPNet \cite{fu2019lightweight}, (f) NPE \cite{wang2013naturalness}, (g) GCANet \cite{chen2019gated}, (h) DualGCN \cite{fu2021rain}, (i) KinD \cite{zhang2019kindling}, (j) TSDNet \cite{liu2022deep}, (k) AirNet \cite{li2022all}, (l) LLFlow \cite{wang2022low}, (m) WeatherDiff \cite{ozdenizci2023restoring}, and (n) ERANet.}
		\label{Figure_real}
	\end{figure*}
\subsection{Referenced Low-Visibility Enhancement Analysis}
   In this subsection, ERANet and competitive methods are used to enhance three types of low-visibility (i.e., hazy, rainy, and low-light) images. The quantitative and qualitative analysis will also be exploited to evaluate the enhancement results.
\subsubsection{Dehazing}
   We first compute objective evaluation metrics for the test images from RESIDE-OTS \cite{li2018benchmarking}, Seaships \cite{shao2018seaships} and SMD \cite{prasad2017video}. As shown in Table \ref{Table_Haze}, ERANet generates the best evaluation results from both PSNR and SSIM metrics. Although NIQE does not produce the best results, its performance remains stable in comparison to other methods.
   
   As shown in Fig. \ref{Figure_haze}, the sky and water areas in the DCP-based restored images are distorted, leading to unnatural color performance. Due to the apparent difference between the inverted hazy and low-light images, SDD has difficulty in accurately removing the haze effects. The ROP+ also fails to generate satisfactory imaging results, since the image contrast is low and the color is off-white. The dehazing results, yielded by AirNet, suffer from unnatural black artifacts, leading to the severely visually degraded images. In contrast, MIRNetv2 performs well when the haze concentration is low, but it fails to accurately extract the potential features from the dense haze. The robustness of TransWeather is extremely sensitive to different haze concentrations, leading to unstable dehazing results under different imaging conditions. WeatherDiff generates unnatural dehazing results during the process of generating haze-free images. The visually-degraded scenes will bring negative influences on higher-level computer vision tasks, e.g., object detection, recognition, tracking and segmentation, etc. Compared with other methods, our ERANet is capable of successfully enhancing hazy images with better detail preservation and color recovery performance.
\subsubsection{Deraining}
   Similar to the above dehazing experiments, several objective evaluation metrics will be exploited for the test images from the Rain100L \cite{yang2019joint}, Seaships \cite{shao2018seaships} and SMD \cite{prasad2017video}, shown in Table \ref{Table_Rain}. The KRM of our ERANet can provide sufficient information on edge prior, lessen its reliance on network learning, and generalize scenes more effectively. Therefore, ERANet has the capacity of achieving satisfactory deraining results at a low computational cost.
   
   The visual effects of different deraining methods are shown in Fig. \ref{Figure_rain}. DualGCN could reconstruct satisfactory visual versions from the quality-degraded images. However, some rain streaks are still noticeable, leading to visual image degradation with partially-local image regions. AirNet is also susceptible to incomplete rain removal when it is intractable to distinguish the rain streaks and background. MIRNetv2 fails to separate unwanted rain streaks from the rainy images, primarily due to the strong dependence of deep networks on training data. TranWeather could eliminate most of the rain streaks, but the local image regions are still negatively affected by the residual rain streaks. KBNet can distinguish the unwanted rain streaks and complex backgrounds when the rain streaks are not visually prominent. WeatherDiff easily introduces unnaturally blocky shadows in local areas and exhibits incomplete rainy effect removal, resulting in low-quality restored images. Our ERANet performs well in handling complicated rainy artifacts under different imaging scenarios. The corresponding natural-looking appearance seems to be more similar to the ground-truth version.
\subsubsection{Low-Light Image Enhancement}
   As shown in Table \ref{Table_Low}, several objective evaluation metrics are also exploited to evaluate the low-light image enhancement results. The test images are directly extracted from the LOL \cite{wei2018deep}, Seaships \cite{shao2018seaships} and SMD \cite{prasad2017video}. Our ERANet could generate the best quantitative results under consideration in most of the cases. The color and edge information of low-light images are often hidden in the dark regions, easily resulting in color distortion and loss of edge textures in the enhanced images. Benefiting from the edge detection operators, ERANet has the capacity of accurately extracting the edge features, and effectively preserving the image color and textural details, etc.
   
   Fig. \ref{Figure_low} visually displays the enhanced images yielded by different low-light enhancement methods. SDD fails to generate satisfactory enhancement results, whose appearances are similar to the original degraded images. It is difficult to exploit ROP+ to handle the complex low-light imaging scenarios. MIRNetv2 struggles to extract the potential feature information from dark backgrounds and exhibits color distortion in local areas. Both TransWeather and SMNet perform poorly on the SMD dataset, mainly because the collected images contain large sea (i.e., water surface) and sky regions. WeatherDiff still exhibits unnatural black patches in local areas, leading to visual quality degradation under different imaging scenes. Compared with these imaging methods, our ERANet achieves a better balance between luminance enhancement and detail preservation.
\subsubsection{Low-Visibility Enhancement on Standard Datasets}
   To evaluate the generalization ability of ERANet for different low-visibility scenes, we selected three standard datasets, i.e., RESIDE-OTS for dehazing \cite{li2018benchmarking}, Rain100L for deraining \cite{yang2019joint}, and LOL for low-light image enhancement \cite{wei2018deep}.
   
   Fig. \ref{Figure07_SD} visually displays the multi-scene visibility enhancement results under different weather conditions. Our ERANet is compared with several state-of-the-art imaging methods, i.e., ROP+ \cite{liu2022rank}, LPNet \cite{fu2019lightweight}, NPE \cite{wang2013naturalness}, GCANet \cite{chen2019gated}, DualGCN \cite{fu2021rain}, KinD \cite{zhang2019kindling}, TSDNet \cite{liu2022deep}, AirNet \cite{li2022all}, LLFlow \cite{wang2022low}, and WeatherDiff \cite{ozdenizci2023restoring}. It can be found that ERANet can effectively improve the overall brightness and contrast for hazy and low-light images, and accurately separate the rain streaks from the background for rainy images. The meaningful textures and sharp edges could be adequately reconstructed, leading to visual quality improvement. For the LOL dataset, which is the first real-world benchmark containing paired normal/low-light images, ERANet can still generate satisfactory enhanced images whose intensities are the closest to the real values. In addition, compared with other competitive imaging methods, it yields more robust visibility enhancement results under different experimental scenarios. These experiments have verified the powerful generalization ability of ERANet for multi-scene visibility enhancement under complex weather conditions.
\subsection{No-Reference Low-Visibility Enhancement Analysis}
   To further demonstrate the superiority of ERANet in practical applications, we also conduct numerous imaging experiments on real-world maritime-related low-visibility images. The high-quality enhancement performance is beneficial for accurately detecting or segmenting the surface objects of interest. It can provide useful perceptual information for promoting the navigational safety of vessels under complex weather conditions. As shown in Fig. \ref{Figure_real}, our ERANet is compared with $10$ state-of-the-art imaging methods, which perform well in synthetic experiments, for subjective visual analysis. It is observed that our method is capable of reconstructing the structural features from the quality-degraded images. In contrast, other competing methods generate the restored images, which easily suffer from appearance and geometric distortion. For complex weather in realistic navigational environments, ERANet has the capacity of implementing no-reference low-visibility enhancement with high robustness and effectiveness. The reliable results are useful for marine surface vessels to guarantee safety in waterborne transportation systems.
	\setlength{\tabcolsep}{10.5pt}
	\begin{table}[t]
		%\scriptsize
		\centering
		\caption{Ablation study of our ERANet based on the combination of CAM, SAM, and KRM on Rain100L dataset \cite{yang2019joint}.}
		\begin{tabular}{ccc|cc}
			\hline
			CAM             & SAM            & KRM            & {PSNR $\uparrow$}  & SSIM $\uparrow$  \\ \hline \hline
			                &                &                & 32.13$\pm$2.89     & 0.937$\pm$0.031  \\
			  \CheckmarkBold  &                &                & 34.31$\pm$2.71     & 0.958$\pm$0.026  \\
			                & \CheckmarkBold &                & 34.34$\pm$2.88     & 0.961$\pm$0.029  \\
			                &                & \CheckmarkBold & 35.21$\pm$3.55     & 0.963$\pm$0.027  \\
			  \CheckmarkBold  & \CheckmarkBold &                & 35.33$\pm$3.29     & 0.965$\pm$0.031  \\
			  \CheckmarkBold  &                & \CheckmarkBold & 35.14$\pm$3.42     & 0.966$\pm$0.033  \\
			                & \CheckmarkBold & \CheckmarkBold & 35.23$\pm$3.29     & 0.965$\pm$0.029  \\
			  \CheckmarkBold  & \CheckmarkBold & \CheckmarkBold & 35.78$\pm$3.54     & 0.970$\pm$0.025  \\ \hline
		\end{tabular}
		\label{Table_ablation_earb}
	\end{table}
	\setlength{\tabcolsep}{21.0pt}
	\begin{table}[t]
		\centering
		\caption{Ablation study of the different types of edge detection operators on Rain100L dataset \cite{yang2019joint}.}
		\begin{tabular}{l|cc}
			\hline
			Operator    & PSNR $\uparrow$  & SSIM $\uparrow$     \\ \hline \hline
			---         & 32.33$\pm$5.14   & 0.914$\pm$0.053     \\			
			Roberts     & 33.55$\pm$4.97   & 0.933$\pm$0.041     \\
			Prewitt     & 33.62$\pm$4.21   & 0.935$\pm$0.039     \\
			Sobel       & 34.91$\pm$4.08   & 0.945$\pm$0.036     \\
			Laplacian   & 35.27$\pm$3.37   & 0.953$\pm$0.031     \\ \hline
			Kirsch      & 35.78$\pm$3.54   & 0.970$\pm$0.025     \\ \hline
		\end{tabular}
		\label{Table_ablation_oper}
	\end{table}
\setlength{\tabcolsep}{10.0pt}
	\begin{table}[t]
		%\scriptsize
		\centering
		\caption{Ablation study of the proposed loss function on Rain100L dataset \cite{yang2019joint}.}
		\begin{tabular}{ccc|cc}
			\hline
			$\mathcal{L}_{\emph{\text{MS-SSIM}}}$  & $\mathcal{L}_{\ell_1}$      & $\mathcal{L}_{TV}$     & {PSNR $\uparrow$} & SSIM $\uparrow$  \\ \hline \hline
			\CheckmarkBold &                &                 & 35.11$\pm$2.78                       & 0.957$\pm$0.027                                       \\
			                & \CheckmarkBold &                & 34.75$\pm$2.94                       & 0.951$\pm$0.026                                       \\
			\CheckmarkBold & \CheckmarkBold &                 & 35.47$\pm$3.22                       & 0.961$\pm$0.027                                         \\
			\CheckmarkBold & \CheckmarkBold & \CheckmarkBold  & 35.78$\pm$3.54                       & 0.970$\pm$0.025                                          \\ \hline
		\end{tabular}
		\label{Table_ablation_loss}
	\end{table}
	\begin{figure}[t]
		\centering
		\includegraphics[width=1.00\linewidth]{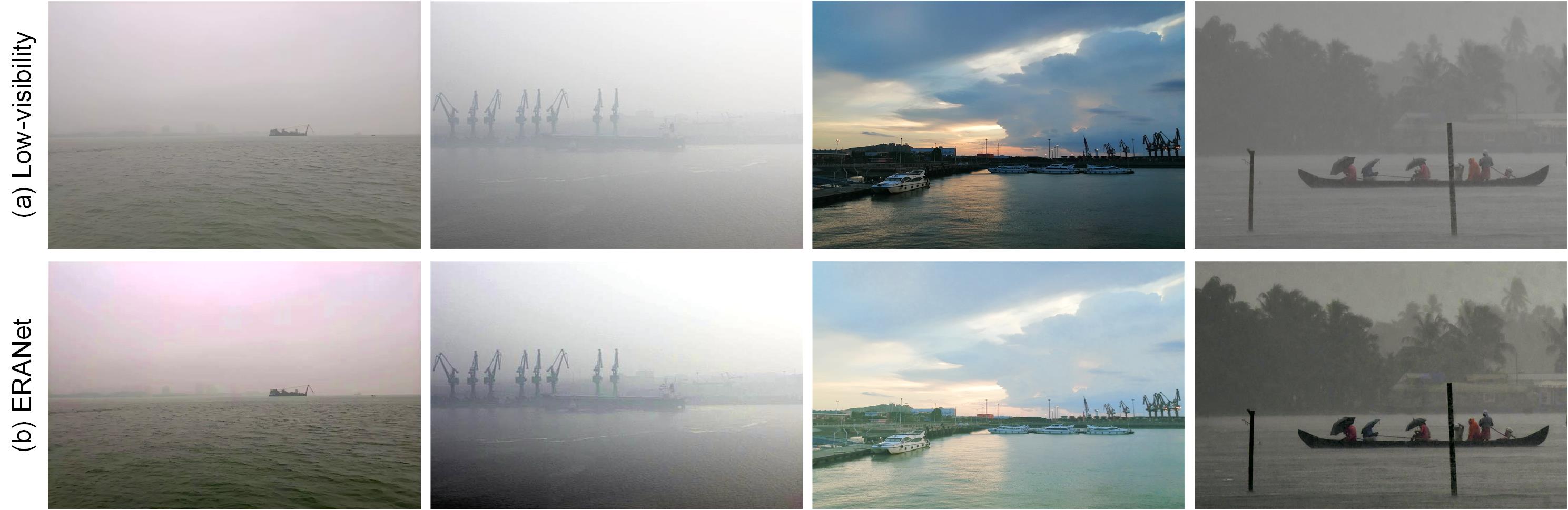}
		\caption{Some examples of failure cases for our ERANet. From top to bottom: (a) real-world low-visibility images, and (b) ERANet-generated results.}
		\label{Figure_failure}
	\end{figure}
\subsection{Ablation Study}\label{as}
   The ablation study is a valuable method to investigate which modules play more important roles in network learning. Compared with dehazing and low-light enhancement, draining seems more challenging due to the structural degradation, severe occlusion, and complex composition, etc. Therefore, we only exploit the draining experiments to perform the ablation study from three aspects, i.e., edge-guided attention residual block, edge detection operators, and loss function.
\subsubsection{Edge-Guided Attention Residual Block}
   We conduct numerous experiments to verify the rationality of elaborately designed parts in the edge-guided attention residual block (EARB). As shown in Table \ref{Table_ablation_earb}, the imaging performance will be noticeably worse if EARB only consists of basic residual blocks without additional modules to assist parameter learning. The quantitative evaluation results could be significantly improved due to the incorporation of KRM, which is capable of extracting meaningful edge features. The combination of CAM and SAM enables the improvement of image quality but still fails to effectively remove rain streaks and raindrops, leading to unsatisfactory evaluation results. The combination of CAM, SAM, and KRM can generate the most satisfactory imaging results, demonstrating that the attention mechanism and structural reparameterization play significant roles in our edge-guided attention residual block.
\subsubsection{Edge Detection Operators}
   This subsection will compare the benefits of Kirsch and the other four edge detection operators (i.e., Roberts, Prewitt, Sobel, and Laplacian) for the reparameterization module. In particular, the Roberts, Prewitt, and Sobel are typical first-order differential operators, which compute the gradients in both vertical and horizontal directions. The Laplacian operator is a second-order differential operator, which can detect the positions and directions of image edges in the insensitivity to random noise. In this work, we individually incorporate the Roberts, Prewitt, Sobel, Laplacian, and Kirsch operators into the reparameterization module for retraining and objective analysis. The results of ablation studies are illustrated in Table \ref{Table_ablation_oper}. It can be found that the Kirsch operator generates the best quantitative evaluation results (i.e., PSNR and SSIM) since it can adequately extract the gradient features in all eight directions. Therefore, the Kirsch operator is beneficial for effectively removing the rain streaks, even though the distributions of rain streaks vary complicatedly in different local areas. The high-quality images can be accordingly guaranteed to promote the navigational safety of vessels under complex weather conditions.
\subsubsection{Loss Function Analysis}
   Different sub-loss functions have diverse characteristics, which bring different effects on network training and performance. The selection of a proper loss function highly depends on the specific requirements and task characteristics. The influences of different combinations of sub-loss functions on draining are shown in Table \ref{Table_ablation_loss}. Both PSNR and SSIM are jointly exploited to quantitatively evaluate the imaging performance. It is obvious that the combination of all three sub-loss functions (i.e., $\mathcal{L}_{\emph{\text{MS-SSIM}}}$, $\mathcal{L}_{\ell_1}$ and $\mathcal{L}_{TV}$) yields the highest PSNR and SSIM values. The other combinations or individual utilization bring negative effects on the evaluation results. This is mainly because each sub-loss function has its advantages. The whole combination can take full use of the different strengths, leading to a proper balance between visibility enhancement and edge preservation. Therefore, we propose to jointly employ $\mathcal{L}_{\emph{\text{MS-SSIM}}}$, $\mathcal{L}_{\ell_1}$ and $\mathcal{L}_{TV}$ to stabilize the network training and high-quality image generation.
\subsection{Failure Cases}\label{as}
    Numerous experiments under different imaging conditions have demonstrated the superior performance of our ERANet. However, it still suffers from some failure cases in practical applications. For example, in Fig. \ref{Figure_failure}, it is challenging to restore a haze-free or normal-light image when the hazy image is excessively dark or when the local area of the low-light image is brighter. This difficulty may stem from the disparate data distribution of these visually-degraded images. In addition, many dim and blurry images have low pixel values and obscured background content, making it difficult to accurately extract the potential features, especially when different types of visual degradation occur simultaneously.
\subsection{Improvement of Object Detection}
   \begin{figure*}[t]
	\centering
         \includegraphics[width=1.00\linewidth]{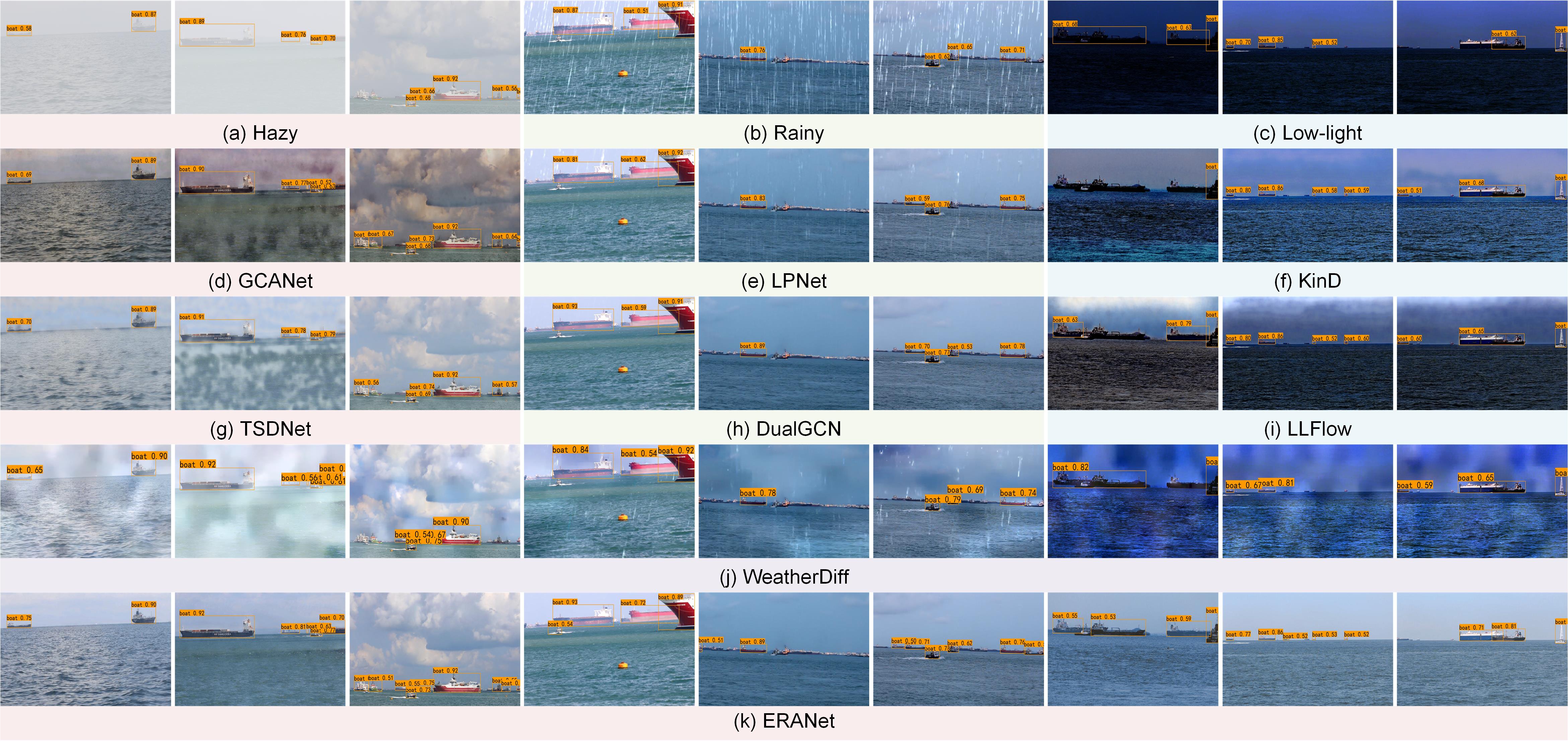}
         \caption{Comparisons of YOLOv7-based object detection results for visually-degraded images and their restored versions. From top to bottom: (a) hazy images, (b) rainy images, (c) low-light images, restored images using (d) GCANet \cite{chen2019gated}, (e) LPNet \cite{fu2019lightweight}, (f) KinD \cite{zhang2019kindling}, (g) TSDNet \cite{liu2022deep}, (h) DualGCN \cite{fu2021rain}, (i) LLFlow \cite{wang2022low}, (j) WeatherDiff \cite{ozdenizci2023restoring}, and (k) ERANet, respectively.}
		\label{Figure_detection}
	\end{figure*}
   To further demonstrate the practical advantages of ERANet in maritime scenarios, we directly exploit the popular YOLOv7 \cite{wang2022yolov7} to detect vessel objects from the original low-visibility images and the enhanced images, which are generated using the GCANet \cite{chen2019gated}, LPNet \cite{fu2019lightweight}, KinD \cite{zhang2019kindling}, TSDNet \cite{liu2022deep}, DualGCN \cite{fu2021rain}, LLFlow \cite{wang2022low}, WeatherDiff \cite{ozdenizci2023restoring}, and our ERANet. The experimental images are extracted from the SMD dataset \cite{prasad2017video}. As shown in Fig. \ref{Figure_detection}, the YOLOv7 detector cannot guarantee accurate object detection in low-visibility scenes due to low contrast and vague edge features. After the restoration of low-visibility images, the detection accuracy could be increased since the enhanced images contain more meaningful features. However, the competing methods easily fail to guarantee high-accuracy detection in extremely low-visibility scenes. This is mainly because the loss of fine details could bring negative effects on object detection. Compared with these imaging methods, our ERANet could generate satisfactory results with higher robustness and accuracy. The superior performance will be more pronounced when the image quality becomes worse. It demonstrates that ERANet is more beneficial for higher-level visual tasks for marine surface vessels under multi-scene low-visibility scenarios.
   \setlength{\tabcolsep}{1.75pt}
   \begin{table}[t]
   %\scriptsize
        \centering
        The comparisons of segmentation performance (mIoU) for different multi-scene visibility enhancement methods under hazy, rainy, and low-light conditions. The popular SMD dataset \cite{prasad2017video} is exploited to quantitatively evaluate the segmentation results. The best three results are highlighted in {\color{red}red}, {\color{blue}blue}, and {\color{green}green} colors, respectively.
        \begin{tabular}{l|cccc}
        \hline
        Methods               & Hazy            & Rainy          & Low-Light            & Average         \\ \hline\hline
        GCANet \cite{chen2019gated}       & 88.62$\pm$7.48  & --- & --- & --- \\ 
        TSDNet \cite{liu2022deep}       & \color{green}{91.24$\pm$7.10}  & --- & --- & --- \\ 
        LPNet \cite{fu2019lightweight}       & ---  & \color{green}{85.64$\pm$10.41} & --- & --- \\ 
        DualGCN \cite{fu2021rain}       & ---  & \color{blue}{90.50$\pm$5.88} & --- & --- \\ 
        KinD \cite{zhang2019kindling}       & ---  & --- & \color{green}{89.94$\pm$7.28} & --- \\ 
        LLFlow \cite{wang2022low}       & ---  & --- & 89.48$\pm$6.51 & --- \\ \hline
        MIRNetv2 \cite{zamir2022learning}       & \color{red}{91.57$\pm$7.56}  & 71.61$\pm$10.11 & 89.21$\pm$7.53 & \color{blue}{84.13$\pm$12.30} \\ 
        TransWeather \cite{valanarasu2022transweather}   & 86.81$\pm$8.79  & 72.96$\pm$8.51  & \color{blue}{90.24$\pm$5.34} & \color{green}{83.34$\pm$10.73} \\ 
        WeatherDiff \cite{ozdenizci2023restoring}    & 81.10$\pm$5.92  & 76.66$\pm$6.34  & 80.10$\pm$8.23 & 79.29$\pm$7.16  \\ \hline
        ERANet         & \color{blue}{91.54$\pm$4.77}  & \color{red}{92.15$\pm$4.04}  & \color{red}{93.62$\pm$4.90} & \color{red}{92.44$\pm$4.67}  \\ \hline
        \end{tabular}
    \label{Table:scene}
    \end{table}
\subsection{Improvement of Scene Segmentation}
\begin{figure}[t]
        \centering
        \includegraphics[width=1.00\linewidth]{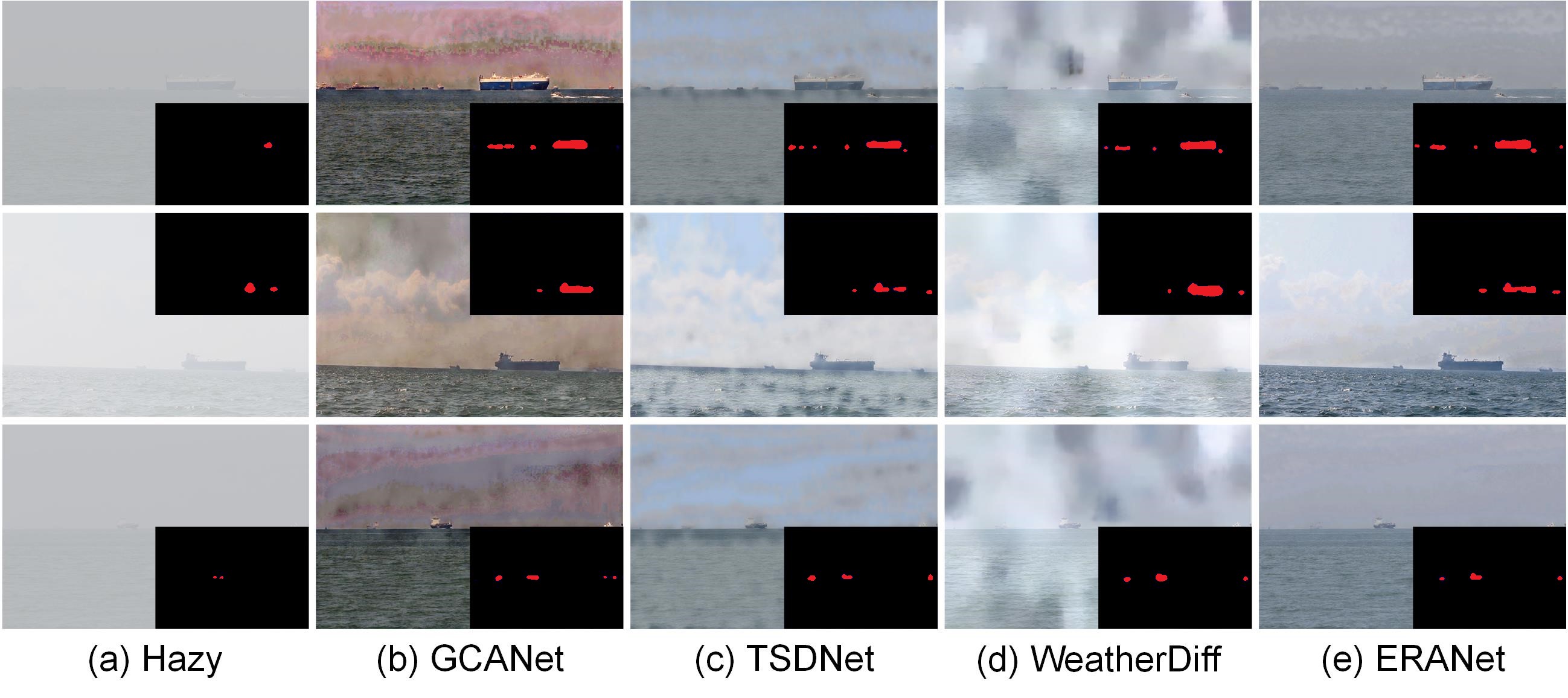}
        \caption{Comparisons of DeepLabv3+-based scene segmentation results for hazy images and their restored versions. From left to right: (a) hazy images \cite{prasad2017video}, restored images using (b) GCANet \cite{chen2019gated}, (c) TSDNet \cite{liu2022deep}, (d) WeatherDiff \cite{ozdenizci2023restoring}, and (e) ERANet, respectively.}
		\label{Figure_segmentation_Haze}
  \end{figure}

\begin{figure}[t]
	\centering
        \includegraphics[width=1.00\linewidth]{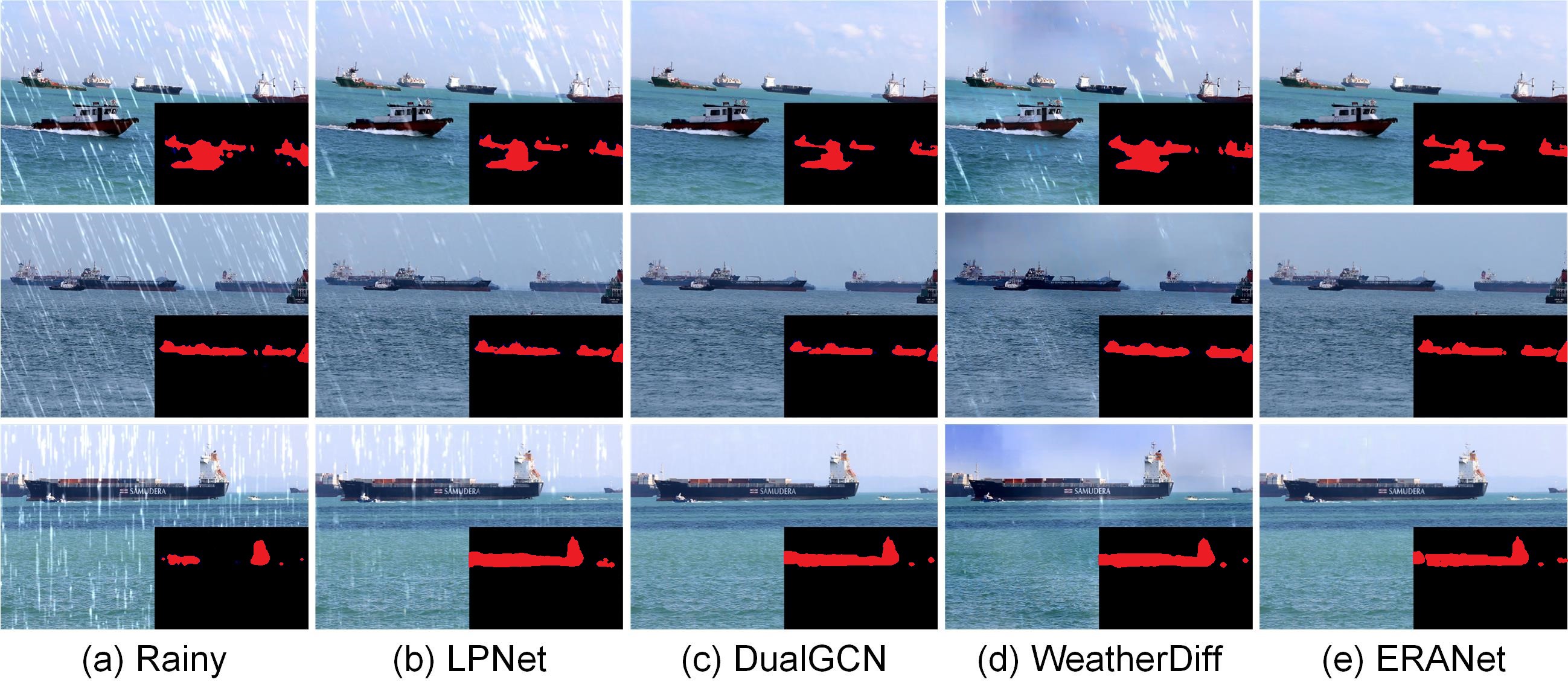}
        \caption{Comparisons of DeepLabv3+-based scene segmentation results for rainy images and their restored versions. From left to right: (a) rainy images, restored images using (b) LPNet \cite{fu2019lightweight}, (c) DualGCN \cite{fu2021rain}, (d) WeatherDiff \cite{ozdenizci2023restoring}, and (e) ERANet, respectively.}
		\label{Figure_segmentation_Rain}
  \end{figure}

\begin{figure}[t]
	\centering
	\includegraphics[width=1.00\linewidth]{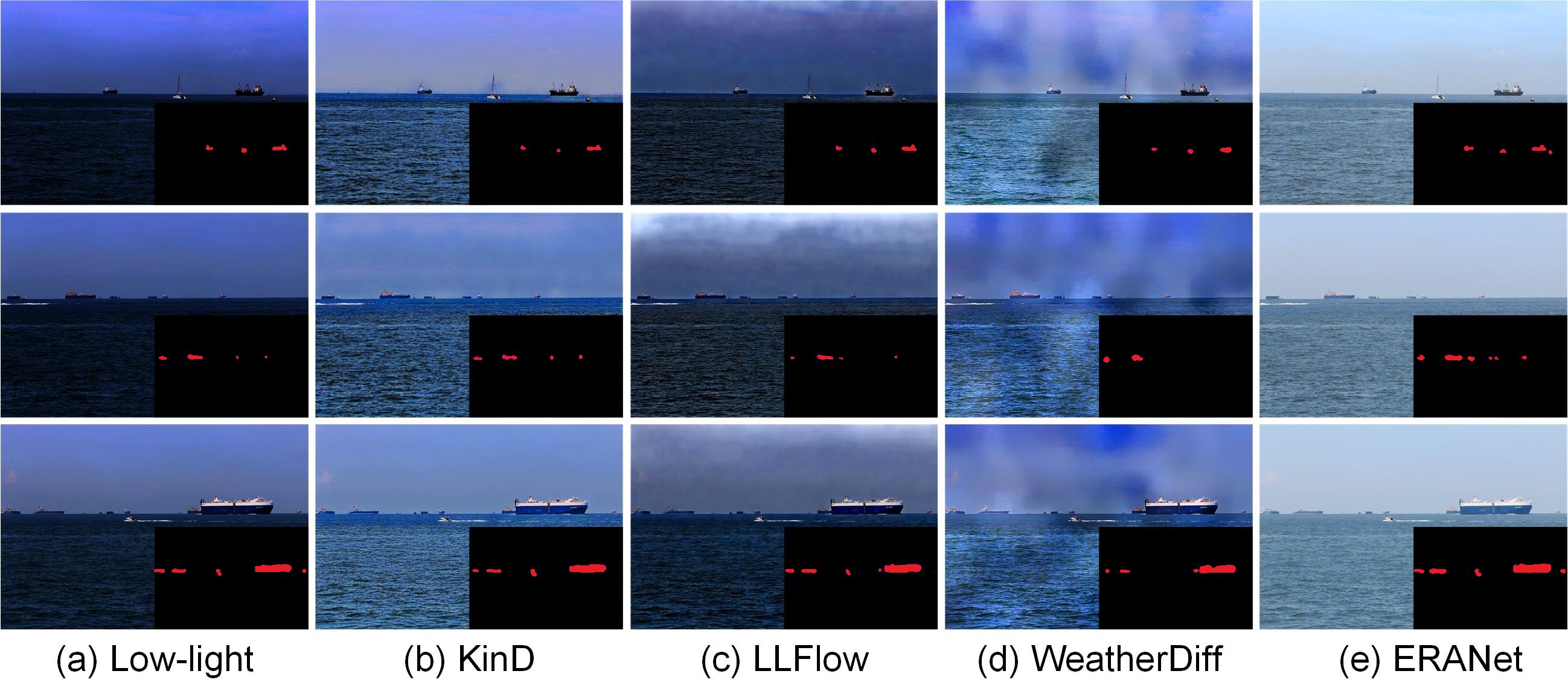}
	\caption{Comparisons of DeepLabv3+-based scene segmentation results for low-light images and their restored versions. From left to right: (a) low-light images, restored images using (b) KinD \cite{zhang2019kindling}, (c) LLFlow \cite{wang2022low}, (d) WeatherDiff \cite{ozdenizci2023restoring}, and (e) ERANet, respectively.}
	\label{Figure_segmentation_Low}
 \end{figure}

   The scene segmentation is also a typical higher-lever visual task, which is conducted on different visibility enhancement results to verify the superiority of our ERANet. The popular DeepLabv3+ \cite{chen2018encoder}, an encoder-decoder structure, is considered as the basic segmentation method. In particular, we directly exploit the officially provided pre-trained model and select several test images from the SMD dataset \cite{prasad2017video}. The mean intersection over union (mIoU) is exploited to quantitatively evaluate the segmentation performance. Table \ref{Table:scene} illustrates the quantitative results on scene segmentation for several multi-scene visibility enhancement methods under different degradations. Our ERANet is capable of generating more robust and reliable enhancement performance under complex weather conditions. It is thus tractable to address the ship collision avoidance with the corresponding higher-quality segmentation results. The visual segmentation results are shown in Figs. \ref{Figure_segmentation_Haze}-\ref{Figure_segmentation_Low}. The edge features of objects in low-visibility environments are blurry and low-contrast, which make the scene segmentation challenging to accurately classify the pixels. However, due to the noise interference and color distortion brought by other competing methods, the corresponding segmentation results suffer from over-segmentation or mistakenly classify some parts of the background as objects. After the implementation of ERANet, higher-quality enhanced images could be obtained with better color naturalness and more visible features. Therefore, the challenging pixels could be accurately classified in the scene segmentation results. It will benefit marine surface vessels to detect the navigable waterways to improve navigational safety under complex weather conditions.
\subsection{Comparison of Running Time}
   To further evaluate the imaging efficiency, shown in Table \ref{Table_time}, our EARNet is compared with several representative visibility enhancement methods in terms of model size and running time. All competing methods considered in this work will run and calculate the running time under PC with Intel(R) Core(TM) i9-12900K CPU @2.30GHz and Nvidia GeForce RTX 3080 Ti Laptop GPU. The collected images with the resolution of $1920 \times 1080$ pixels (i.e., $1080$p) are adopted in our numerical experiments. With superior enhancement performance, our method achieves $1080$p scene recovery over $40$fps on the experimental platform, which is faster than most previous methods. It is thus flexible and feasible to incorporate the EARNet into the onboard sensors and computational devices for marine surface vessels in IWTS.
   \setlength{\tabcolsep}{3.0pt}
       \begin{table}[t]
       \centering
       \caption{Comparison of the model size and running time between ERANet and other methods of the $1080$p image ($1920\times1080$ pixels).}
       \begin{tabular}{l|cccc}
		\hline
		Methods                                        & Language   & Frame      & Model Size (KB) & Time (s) \\ \hline \hline
			DCP \cite{he2010single}                        & Matlab (C) & ---        & ---             & 3.211    \\
			NPE \cite{wang2013naturalness}                 & Matlab (C) & ---        & ---             & 23.146   \\
			SDD \cite{wang2013naturalness}                 & Matlab (C) & ---        & ---             & 15.074   \\  
			ROP+ \cite{liu2022rank}                        & Matlab (C) & ---        & ---             & 1.503    \\  \hline
			DDN \cite{fu2017removing}                      & Python     & Tensorflow & 228             & 0.867    \\
			RetinexNet \cite{wei2018deep}                  & Python     & Tensorflow & 1738            & 1.898    \\
			KinD \cite{zhang2019kindling}                  & Python     & Tensorflow & 4014            & 1.748    \\
			LPNet \cite{fu2019lightweight}                 & Python     & Tensorflow & 1513            & 0.592    \\		
			GCANet \cite{chen2019gated}                    & Python     & Pytorch    & 2758            & 0.148    \\
			DIG \cite{ran2020single}                       & Matlab (C) & ---        & ---             & 2.452    \\
			DualGCN \cite{fu2021rain}                      & Python     & Tensorflow & 10669           & 22.976   \\        
			LLFlow \cite{wang2022low}                      & Python     & Pytorch    & 21362           & 0.301    \\        		
			TSDNet \cite{liu2022deep}                      & Python     & Pytorch    & 14275           & 0.032    \\ 
			AirNet \cite{li2022all}                        & Python     & Pytorch    & 35407           & 1.725    \\ 
			MIRNetv2 \cite{zamir2022learning}              & Python     & Pytorch    & 23006           &3.562     \\ 
			TransWeather \cite{valanarasu2022transweather} &Python      & Pytorch    & 85669           & 1.990    \\ 
			SMNet \cite{lin2023smnet}                      &Python      & Pytorch    & 11880           & 0.013    \\ 
			KBNet \cite{zhang2023kbnet}                    & Python     & Pytorch    & 115887          & 4.189    \\ 
			USCFormer \cite{wang2023uscformer}             & Python     & Pytorch    & 63406           & 2.025    \\ 
			WeatherDiff \cite{ozdenizci2023restoring}      & Python     & Pytorch    & 1296804         & 31.342   \\ \hline
			ERANet                                         & Python     & Pytorch    & 2449            & 0.016    \\ \hline
	\end{tabular}
    \label{Table_time}
 \end{table}
\section{Conclusions and Future Perspectives}\label{conc}
   This work proposes an edge reparameterization- and attention-guided network (ERANet), which is essentially a \textit{general-purpose} multi-scene visibility enhancement method. It can real-timely recover low-visibility scenes using only one network under different weather conditions. In particular, we design an edge-guided attention residual block, motivated by the Kirsch-guided reparameterization module, which enables ERANet to improve the visual perception of low-visibility scenes with low computational cost. The comprehensive experiments on standard and IWTS-related datasets have demonstrated that ERANet is comparable or superior to state-of-the-art visibility enhancement methods on several quantitative metrics. In addition, according to the experimental results on object detection and scene segmentation, our ERANet could make a major contribution toward higher-level computer vision tasks under low-visibility scenes in IWTS. To make visibility enhancement more reliable and applicable, we will further extend the related work along with the following directions.
\begin{itemize}
    \item The current ERANet only performs well in parameter learning and reasoning at a single scale. In contrast, the image edges essentially have different widths and characteristics at different scales. However, the Kirsch operators are fixed and cannot adaptively adjust the multi-scale features. Therefore, we will further focus on how to better learn the multi-scale features \cite{he2015spatial,li2021multi} without excessively increasing the computing burden of edge devices.
    \item Numerous efforts have been devoted to other low-level vision tasks (e.g., image desnowing \cite{kulkarni2022wipernet,ding2023cf} and image super-resolution \cite{chen2023dual,zhou2023srformer}) in intelligent transportation systems. To achieve more flexible and feasible imaging results under more weather conditions and different task requirements, the recovery and reconstruction capabilities for more imaging scenes will be incorporated into our ERANet-based visibility enhancement framework.
    \item The higher-level computer vision tasks (i.e., object detection and scene segmentation) are performed after the implementation of visibility enhancement in this work. This two-step strategy easily suffers from complicated computations in practical applications. Motivated by the multi-task learning (MTL) \cite{zhang2018overview}, it is necessary to simultaneously execute the tasks of multi-scene visibility enhancement and higher-level computer vision. The corresponding visual computing process could thus be simplified and more portable in IWTS.
\end{itemize}

Benefiting from the incorporated attention mechanism and structural heavy-parameter modules, ERANet has the capacity of real-timely enhancing various types of low-visibility images using only one network. Therefore, there exists great potential in the application of ERANet for promoting the navigational safety of moving vessels under complex weather conditions.
%
%
%
%\ifCLASSOPTIONcaptionsoff
%	\newpage
%	\fi
%	\bibliographystyle{IEEEtran}
%	\bibliography{ref.bib}
%	\tiny	
	%
	%
\bibliography{ref}    
\bibliographystyle{IEEEtran}
\end{document}